\documentclass[runningheads]{llncs}
\usepackage{graphicx}

\usepackage{tikz}
\usepackage{comment}
\usepackage{amsmath,amssymb} 
\usepackage{color}

\usepackage[accsupp]{axessibility}  

\usepackage{booktabs}
\usepackage{multirow}
\usepackage{algorithm}
\usepackage{algpseudocode}
\usepackage{pifont}
\usepackage{caption}
\usepackage{subcaption}

\usepackage{array}
\newcommand{\PreserveBackslash}[1]{\let\temp=\\#1\let\\=\temp}
\newcolumntype{C}[1]{>{\PreserveBackslash\centering}p{#1}}
\newcolumntype{R}[1]{>{\PreserveBackslash\raggedleft}p{#1}}
\newcolumntype{L}[1]{>{\PreserveBackslash\raggedright}p{#1}}

\begin{document}

\newcommand*{\argmax}{arg\,max}
\newcommand*{\argmin}{arg\,min}
\def\eg{\emph{e.g.}} 
\def\Eg{\emph{E.g.}}
\def\ie{\emph{i.e.}} 
\def\Ie{\emph{I.e.}}
\def\cf{\emph{c.f.}} 
\def\Cf{\emph{C.f.}}
\def\etc{\emph{etc.}} 
\def\vs{\emph{v.s.}}
\def\wrt{w.r.t.} 
\def\dof{d.o.f.}
\def\etal{\emph{et al.}}

\makeatletter
\newcommand{\printfnsymbol}[1]{%
  \textsuperscript{\@fnsymbol{#1}}%
}

\pagestyle{headings}
\mainmatter
\def\ECCVSubNumber{2124}  

\title{GeoRefine: Self-Supervised Online Depth Refinement for Accurate Dense Mapping} 

\titlerunning{GeoRefine}
%
\author{Pan Ji\thanks{Joint first authorship. P. Ji is the corresponding author (peterji530@gmail.com).}, Qingan Yan\printfnsymbol{1}, Yuxin Ma, and Yi Xu 
}
\institute{OPPO US Research Center, InnoPeak Technology}
\titlerunning{GeoRefine}
\authorrunning{P. Ji \etal}

\maketitle

\begin{center}
\vspace{-0.3cm}
\captionsetup{type=figure}
\includegraphics[width=0.95\textwidth]{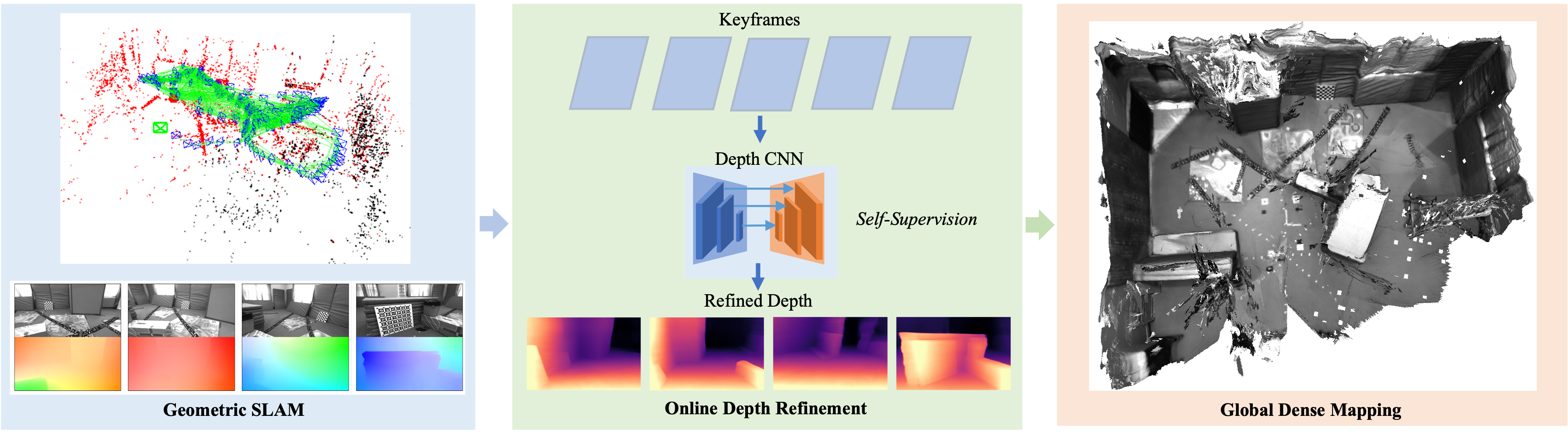}
\captionof{figure}{We present an online depth refinement system for geometrically-consistent dense mapping from monocular data. Our system starts with geometric SLAM that is made robust by incorporating learning-based priors. Together with map points and camera poses from SLAM, a depth CNN is continuously updated using self-supervised losses. A globally consistent map is finally reconstructed from refined depth maps via an off-the-shell TSDF fusion method.}
\label{fig:teaser}
\vspace{-0.3cm}
\end{center}

\begin{abstract}
    We present a robust and accurate depth refinement system, named GeoRefine, for geometrically-consistent dense mapping from monocular sequences. 
    GeoRefine consists of three modules: a hybrid SLAM module using learning-based priors, an online depth refinement module leveraging self-supervision, and a global mapping module via TSDF fusion.
    The proposed system is online by design and achieves great robustness and accuracy via: (i) a robustified hybrid SLAM that incorporates learning-based optical flow and/or depth; (ii) self-supervised losses that leverage SLAM outputs and enforce long-term geometric consistency; (iii) careful system design that avoids degenerate cases in online depth refinement. We extensively evaluate GeoRefine on multiple public datasets and reach as low as $5\%$ absolute relative depth errors.
\end{abstract}

\section{Introduction}
\label{sec:intro}
3D reconstruction from monocular images has been an active research topic in computer vision for decades~\cite{hartley2003multiple}. Traditionally, a scene is usually reconstructed in the form of a set of {\it sparse} 3D points via geometric techniques, such as Structure-from-Motion (SfM) or Simultaneous Localization and Mapping (SLAM). Over the years, those monocular geometric methods have been continuously improved and become very accurate in recovering 3D map points. Representative open-source systems along this line include COLMAP~\cite{schonberger2016structure} -- an offline SfM system, and ORB-SLAM~\cite{mur2015orb,mur2017orb,campos2020orb} -- an online SLAM system.

Recently, deep-learning-based methods~\cite{eigen2014depth,garg2016unsupervised,godard2019digging} have achieved impressive results in predicting a {\it dense} depth map from a single image. Those models are either trained in a supervised manner~\cite{eigen2014depth,ranftl2019towards,ranftl2021vision} using ground-truth depths, or through a self-supervised framework~\cite{garg2016unsupervised,godard2019digging} leveraging the photometric consistency between stereo and/or monocular image pairs. During inference, with the prior knowledge learned from data, the depth models can generate {\it dense} depth images even in textureless regions. However, the errors in the predicted depths are still relatively high.

A few methods~\cite{tiwari2020pseudo,luo2020consistent} aim to get the best of geometric systems and deep methods. Tiwari~\etal~\cite{tiwari2020pseudo} let monocular SLAM and learning-based depth prediction form a self-improving loop to improve the performance of each module. Luo~\etal~\cite{luo2020consistent} adopt a test-time fine tuning strategy to enforce geometric consistency using outputs from COLMAP.
Nonetheless, both methods require to pre-compute and store sparse map points and camera poses from SfM or SLAM in an offline manner, which is not applicable to many applications where data pre-processing is not possible. For example, after we deploy an agent to an environment, we want it to automatically improve its 3D perception capability as it moves around. In such a scenario, an online learning method is more desirable.

In this paper, we propose to combine geometric SLAM and a single-image depth model within an {\bf online} learning scheme (see Fig.~\ref{fig:teaser}). The depth model can be any model that has been pretrained either with a self-supervised method~\cite{godard2019digging} or in a supervised fashion~\cite{ranftl2019towards,ranftl2021vision}. Our goal is then to incrementally refine this depth model on the test sequences in an online manner to achieve geometrically consistent depth predictions over the entire image sequence. Note that SLAM in itself is an online system that perfectly fits our online learning framework, but on the other hand, front-end tracking of SLAM often fails under challenging conditions (\eg, with fast motion and large rotation). To facilitate a robust system, we propose to enhance the robustness of geometric SLAM with learning-based priors, \eg, RAFT-flow~\cite{teed2020raft}, which has been shown to be both robust and accurate in a wide range of {\it unseen} scenes~\cite{kopf2021robust,zhang2021consistent,teed2021droid}. We then design a parallel depth refinement module that optimizes the neural weights of depth CNN with self-supervised losses. We perform a careful analysis of failure cases of self-supervised refinement and propose a simple yet effective keyframe mechanism to make sure that no refinement step worsens depth results. We further propose a novel occlusion-aware depth consistency loss to promote long-term consistency over temporally distant keyframes. We perform detailed ablation study to verify the effectiveness of each new component of our proposed {\it GeoRefine}, and conduct extensive experiments on several public datasets~\cite{Burri25012016,sturm2012benchmark,dai2017scannet,Geiger2013IJRR},
demonstrating state-of-the-art performance in terms of dense mapping from monocular images.

\section{Related Work}
\label{sec:relatedwork}
In this section, we briefly review a few highly-related areas, \ie, geometric visual SLAM, learning-based SLAM, and monocular depth estimation. We also discuss their connections to our method.

\subsection{Geometric Visual SLAM}
SLAM is an online geometric system that reconstructs a 3D map consisting of 3D points and simultaneously localizes camera poses \wrt{} the map~\cite{cadena2016past}. According to the methods used in front-end tracking, SLAM systems can be roughly classified into two categories: (i) direct SLAM~\cite{engel2014lsd,engel2017direct,schubert2018direct}, which directly minimizes the photometric error between adjacent frames and optimizes the geometry using semi-dense measurements; (ii) feature-based (indirect) SLAM~\cite{song2013parallel,mur2015orb,mur2017orb,campos2020orb}, which extracts and tracks a set of sparse feature points and then computes the geometry in the back-end using these sparse measurements. Geometric SLAM systems have become accurate and robust due to a number of techniques developed over the years, including robust motion estimation~\cite{fischler1981random}, keyframe mechanism~\cite{klein2007parallel}, bundle adjustment~\cite{triggs2000bundle}, and pose-graph optimization~\cite{kummerle2011g}. In our work, we build our system upon one of the state-of-the-art feature-based systems, \ie, ORB-SLAM~\cite{campos2020orb}. We use ORB-SLAM because it is open-source, delivers accurate 3D reconstructions, and supports multiple sensor modes.

\subsection{Learning-Based SLAM}

CNN-SLAM~\cite{tateno2017cnn} is a hybrid SLAM system that uses CNN depth to bootstrap back-end optimization of sparse geometric SLAM and helps recover a metric scale for 3D reconstruction. In contrast, DROID-SLAM~\cite{teed2021droid} builds SLAM from scratch with a deep learning framework and achieves unprecedented accuracy in camera poses, but does not have the functionality of dense mapping. TANDEM~\cite{koestler2022tandem} presents a monocular tracking and dense mapping framework that relies on photometric bundle adjustment and a supervised multi-view stereo CNN model. CodeSLAM~\cite{bloesch2018codeslam} is a real-time learning-based SLAM system that optimizes a compact depth code over a conditional variational auto-encoder (VAE) and simultaneously performs dense mapping. DeepFactor~\cite{czarnowski2020deepfactors} extends CodeSLAM by using fully-differentiable factor-graph optimization. CodeMapping~\cite{matsuki2021codemapping} further improves over CodeSLAM via introducing a separate dense mapping thread to ORB-SLAM3~\cite{campos2020orb} and additionally conditioning VAE on sparse map points and reprojection errors. Our system bears the most similarity with CodeMapping in terms of overall functionalities, but is significantly different in system design and far more accurate in dense mapping.

\subsection{Monocular Depth Estimation}
{\bf Supervised depth estimation} methods dominate the early trials~\cite{eigen2014depth,liu2015learning,ummenhofer2017demon,zhou2018deeptam,teed2018deepv2d,qi2018geonet,liu2022planemvs} in this area. Eigen~\etal~\cite{eigen2014depth} propose the first deep learning based method to predict depth maps via a convolutional neural network and introduce a set of depth evaluation metrics that are still widely used today. Liu~\etal~\cite{liu2015learning} formulate depth estimation as a continuous conditional random field (CRF) learning problem. Fu~\etal~\cite{fu2018deep} leverage a deep ordinal regression loss to train the depth network. A few other methods combine depth estimation with additional tasks, \eg, pose estimation~\cite{ummenhofer2017demon,zhou2018deeptam,teed2018deepv2d} and surface normal regression~\cite{qi2018geonet}.

{\bf Self-supervised depth estimation} has recently become popular~\cite{garg2016unsupervised,zhou2017unsupervised,godard2019digging,hermann2020self,li2020deep,shu2020feature,xiong2021self,ruhkamp2021attention,li2021generalizing}.
Garg~\etal~\cite{garg2016unsupervised} are the first to apply the photometric loss between left-right stereo image pairs to train a monocular depth model in an unsupervised/self-supervised way. Zhou~\etal~\cite{zhou2017unsupervised} further introduce a pose network to facilitate using a photometric loss across neighboring temporal images. Later self-supervised methods are proposed to improve the photometric self-supervision.  
Some methods~\cite{yin2018geonet,zou2018dfnet,ranjan2019competitive} leverage an extra flow network to enforce cross-task consistency, while a few others~\cite{godard2017unsupervised,wang2018learning,bian2019unsupervised} employ new loss terms during training.
A notable recent method (Monodepth2) is by Godard~\etal~\cite{godard2019digging} who achieve great improvements via a few thoughtful designs, including a per-pixel minimum photometric loss, an auto-masking strategy, and a multi-scale framework. New network architectures are also introduced to boost the performance. Along this line, Wang~\etal~\cite{wang2019recurrent} and Zou~\etal~\cite{zou2020learning} exploit recurrent networks to in the pose and/or depth networks. Ji~\etal~\cite{ji2021monoindoor} propose a depth factorization module and an iterative residual pose module to improve depth prediction in indoor environments. Our system is theoretically compatible with all those methods in the pretraining stage. 

Instead of using ground-truth depths, some methods~\cite{li2018megadepth,li2019learning,ranftl2019towards,zhao2020towards,tiwari2020pseudo,luo2020consistent} obtain the training depth data from the off-the-shell SfM or SLAM. Li and Snavely~\cite{li2018megadepth} perform 3D reconstruction of Internet photos via geometric SfM~\cite{schonberger2016structure} and then use the reconstructed depths to train a depth network. Li~\etal~\cite{li2019learning} learn the depths of moving people by watching and reconstructing static people. Ranftl~\etal~\cite{ranftl2019towards,ranftl2021vision} improve generalization performance of the depth model by training the depth network with various sources, including ground-truth depths and geometrically reconstructed ones. 
Zhang~\etal~\cite{zhang2021consistent} extend the work of~\cite{luo2020consistent} to handling moving objects by unrolling scene flow prediction. Kopf~\etal~\cite{kopf2021robust} further bypass the need of running COLMAP via the use of deformation splines to estimate camera poses. Most of those methods
require a pre-processing step to compute and store 3D reconstructions. In contrast, 
our system runs in an online manner without the need of performing offline 3D reconstruction.

\section{Method -- GeoRefine}
\label{sec:method}
\begin{figure}[!t]
\begin{center}
\includegraphics[width=0.85\textwidth]{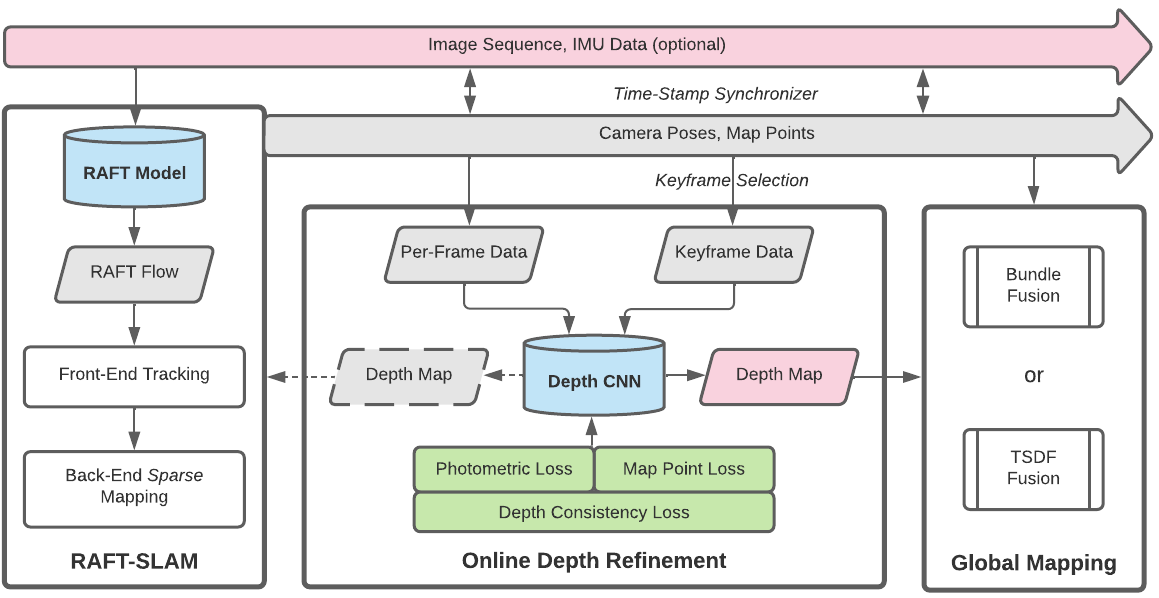}
\end{center}
\vspace{-0.6cm}
\caption{The system workflow of our {\bf GeoRefine}. Our system consists of three main modules, \ie, a RAFT-SLAM module, an Online Depth Refinement module, and a Global Mapping module. Note that keyframe selection in {\it Online Dense Refinement} uses a {\bf different} strategy than in SLAM.}
\label{fig:workflow}
\vspace{-0.5cm}
\end{figure}
In this section, we present {\it GeoRefine}, a self-supervised depth refinement system for geometrically consistent dense mapping from monocular sequences. As shown in Fig.~\ref{fig:workflow}, our system consists of three parallel modules, \ie, a RAFT-SLAM module, an Online Depth Refinement module, and a Global Mapping module. We detail the first two modules in the following sub-sections.

\subsection{RAFT-SLAM}
It is well-known that monocular visual SLAM has several drawbacks: (i) its front-end often fails to track features under adverse environments, \eg, with low-texture, fast motion, and large rotation; (ii) it can only reconstruct the scene up to an {\it unknown} global scale.
To improve the performance of SLAM, a few methods~\cite{tateno2017cnn,yang2018deep,yang2020d3vo} have been proposed to improve back-end optimization of {\it direct} LSD-SLAM~\cite{engel2014lsd}.
In this work, we instead seek to improve the front-end of {\it feature-based} SLAM based on the observation that front-end tracking lose is one of most common causes for SLAM failures and accuracy decrease. We thus present RAFT-SLAM, a hybrid SLAM system that runs a learning-based flow front-end and a traditional back-end optimizer.

\vspace{-0.2cm}
\subsubsection{RAFT-Flow Tracking}\hfill\vspace{0.2cm}\\
RAFT~\cite{teed2020raft} is one of the state-of-the-art flow methods that has shown strong cross-dataset generalization performance. It constructs a correlation volume for all pairs of pixels and uses a gated recurrent unit (GRU) to iteratively update the flow. In our system, we replace the front-end feature matching in ORB-SLAM~\cite{campos2020orb} with RAFT-flow, but still sample sparse points for robust estimation in the back-end. This simple strategy allows us to have the advantages of learning-based flow and traditional robust estimator in one system.

More specifically, for each feature from last frame ${\bf I}_{i-1}$, once it's associated with a map point, we locate its correspondence in incoming frame ${\bf I}_i$ by adding the flow ${\bf F}_{(i-1) \rightarrow i}$. If there are multiple candidates within a predefined radius around a target pixel in ${\bf I}_i$, we choose the one with smallest descriptor residual; or if there is none, we create a new feature instead, with the descriptor being copied from ${\bf I}_{i-1}$. In all our experiments, we set the radius to $1$ pixel. For the sake of robustness, we only keep $N_f = 0.1 \cdot N_t$ matched correspondences for initial pose calculation, where $N_t$ is the total ORB features within the current frame. We note that, compared to leveraging the entire flow, sampling a subset of pixels is more beneficial to the accuracy. We do a forward-backward consistency check on predicted flows to obtain a valid flow mask by using a stringent threshold of $1$ pixel. Similar to \cite{campos2020orb}, we then perform a local map point tracking step to densify potential associations from other views and further optimize the camera pose. The reason why we combine ORB features with flow is that traditional features could help us keep the structure information, mitigating the drifting caused by flow mapping in long sequential tracking. 

\vspace{-0.2cm}
\subsubsection{Multiple Sensor Modes}\hfill\vspace{0.2cm}\\
Our RAFT-SLAM inherits the good properties of ORB-SLAM3~\cite{campos2020orb} in supporting multiple sensor modes. In our system, we consider a minimum sensor setup, \ie, using a monocular camera with (or without) an IMU sensor, so two SLAM modes are under consideration, \ie, the monocular and Visual-Inertial (VI) modes. As we have a CNN depth model to infer the depth map for every image, we additionally form a pseudo-RGBD (pRGBD) mode as in~\cite{tiwari2020pseudo}.

\vspace{-0.3cm}
\noindent\paragraph{\bf Monocular Mode.} Under the monocular mode, RAFT-SLAM reconstructs camera poses and 3D map points in an arbitrary scale. Since we have a pretrained depth model available in our system, we then leverage the CNN predicted depth maps to adapt the scale of map points and camera poses for SLAM. This scale alignment step is necessary in our system because SLAM outputs will be used in the downstream task of refining the depth model. If the scales between these two modules differ too much, depth refinement will be sub-optimal or even totally fail. After initial map points are constructed in our system, we continuously align the scale for a few steps by solving the following least-squares problem:
\begin{equation}
    \min_s \sum_{\bf x} \big(d({\bf x}) - s \cdot \hat{d}({\bf x})\big)^2\;,
\end{equation}
where $s$ is the scale alignment factor to be estimated, and $d({\bf x}), \hat{d}({\bf x})$ are the depth values from a pretrained depth model and SLAM map points respectively. However, if the scales of two modules are already in the same order, \eg, when SLAM runs in the VI or pRGBD model, such an alignment step is not necessary.

\vspace{-0.3cm}
\noindent\paragraph{\bf VI Mode.} VI SLAM is usually more robust than monocular SLAM under challenging environments with low-texture, motion blur and occlusions~\cite{campos2020orb}. Since the inertial sensors provide scale information, camera poses and 3D map points from VI RAFT-SLAM are recovered in metric scale. In this mode, given a scale-aware depth model (\ie, a model that predicts depth in metric scale), we can run the online depth refinement module without taking special care of the scale discrepancies between the two modules.

\vspace{-0.3cm}
\noindent\paragraph{\bf pRGBD Mode.} The pRGBD mode provides a convenient way to incorporate deep depth priors into geometric SLAM. However, we observe that it results in sub-optimal SLAM performance if we naively treat depth predictions as the groundtruth to run the RGBD mode (as done in~\cite{tiwari2020pseudo}) due to noisy predictions. In the RGBD mode of ORB-SLAM3~\cite{campos2020orb}, the depth is mainly used in two SLAM stages, \ie, system initialization and bundle adjustment. By using the input depth, the system can initialize instantly from the first frame, without the need of waiting until having enough temporal baselines. For each detected feature point, employing the depth and camera parameters, the system creates a {\it virtual right correspondence}, which leads to an extra reprojection error term in bundle adjustment~\cite{campos2020orb}. To mitigate the negative impact of the noise in depth predictions, we make two simple yet effective changes in the pRGBD mode as compared to the original RGBD mode: i) we take as input the refined depth maps from the online refinement module (as described in the next subsection) to ensure that the input depth maps are more accurate and temporally consistent; ii) we remove the reprojection error term for the virtual right points in bundle adjustment. 
Note that the input CNN depth is still used in the map point initialization and new keypoint insertion, benefiting the robustness of the SLAM system.
\subsection{Online Depth Refinement}
The depth refinement module receives map points and camera poses from RAFT-SLAM. The depth model is then incrementally refined with self-supervised losses, including a photometric loss, an edge-aware depth smoothness loss, a map-point loss, and a depth consistency loss.

Similar to~\cite{zhou2017unsupervised}, the photometric loss is defined as the difference between a target frame ${\bf I}_i$ and a synthesized frame ${\bf I}_{j\rightarrow i}$ warped from a source frame ${\bf I}_j$ using the depth image ${\bf D}_i$ and the relative pose ${\bf T}_{j\rightarrow i}$, \ie,
\begin{equation}
\label{loss:photo}
    L_p = \sum_j pe({\bf I}_i, {\bf I}_{j\rightarrow i})\;,
\end{equation}
where $pe()$ is the photometric loss function computed with the $\ell_1$ norm and the SSIM~\cite{wang2004image}. Instead of only using 3 neighboring frames to construct the photo-consistency as in~\cite{zhou2017unsupervised,godard2019digging}, we employ a wider baseline photometric loss, \eg, by using a 5-keyframe snippet with $j\in\mathcal{A}_i=\{i-9,i-6,i-3,i+1\}$. Another important difference is that the relative pose ${\bf T}_{j\rightarrow i}$ comes from our RAFT-SLAM, which is more accurate than the one predicted by a pose network. 

Following~\cite{godard2019digging}, we use an edge-aware normalized smoothness loss, \ie,
\begin{equation}
\label{loss:smooth}
    L_s = |\partial_x d_i^*| e^{-|\partial_x I_i|} + |\partial_y d_i^*| e^{-|\partial_y I_i|}\;,
\end{equation}
where $d_i^* = d_i/\Bar{d_i}$ is the mean-normalized inverse depth to prevent depth scale diminishing~\cite{wang2018learning}. 

The map points from RAFT-SLAM have undergone extensive optimization through bundle adjustment~\cite{triggs2000bundle}, so the depths of these map points are usually more accurate than the pretrained CNN depths. As in~\cite{yang2018deep,tiwari2020pseudo}, we also leverage the map-point depths to build a map-point loss as a supervision signal to the depth model. The map-point loss is simply the difference between SLAM map points and the corresponding CNN depths as follows,
\begin{equation}
    \label{loss:mappoint}
    L_{m} = \frac{1}{N_i} \sum_{n=1}^{N_i} \big| {\bf D}_{i, n} - D_{i,n }^{slam} \big|\;,
\end{equation}
where we have $N_i$ 3D map points from RAFT-SLAM after filtering with a stringent criterion (see Sec.~\ref{sec:implementation}) to ensure that only accurate map points are used as supervision. 
In addition to the above loss terms, we propose an occlusion-aware depth consistency loss and a keyframe strategy to build our online depth refinement pipeline.

\vspace{-0.2cm}
\subsubsection{Occlusion-Aware Depth Consistency}\hfill\vspace{0.2cm}\\
Given the depth images of two adjacent images, \ie, ${\bf D}_i$ and ${\bf D}_j$, and their relative pose ${\bf T}=[{\bf R}|{\bf t}]$, we aim to build a robust consistency loss between ${\bf D}_i$ and ${\bf D}_j$ to make the depth predictions consistent with each other. Note that the depth values are not necessarily equal at corresponding positions of frame $i$ and $j$ as the camera can move over time.
With camera pose ${\bf T}$, the depth map ${\bf D}_j$ can be warped and then transformed to a depth map $\Tilde{\bf D}_i$ of frame $i$, via image warping and coordinate system transformation~\cite{bian2019unsupervised,ji2021monoindoor}.
We then define our initial depth consistency loss as,
\begin{equation}
\label{eq:loss_init_consistency}
    L_c({\bf D}_i, {\bf D}_j) = \bigg|1 - {\Tilde{\bf D}_i} / {{\bf D}_i} \bigg|\;.
\end{equation}
However, the loss in Eq.~\eqref{eq:loss_init_consistency} will inevitably include pixels in occluded regions, which hampers model refinement. To effectively handle occlusions, following the per-pixel photometric loss in~\cite{godard2019digging}, we devise a per-pixel depth consistency loss by taking the minimum instead of the average over a set of neighboring frames:
\begin{equation}
\label{loss:consistency}
    L_c = \min_{j\in\mathcal{A}_i} L_c({\bf D}_i, {\bf D}_j)\;.
\end{equation}

\vspace{-0.2cm}
\subsubsection{Degenerate Cases and Keyframe Selection}\hfill\vspace{0.2cm}\\
Self-supervised photometric losses are not without degenerate cases. If they are not carefully considered, self-supervised training or finetuning will be deteriorated, leading to worse depth predictions. A first degenerate case happens when the camera stays {\it static}. This degeneracy has been well considered in the literature. For example, Zhou~\etal~\cite{zhou2017unsupervised} remove static frames in an image sequence by computing and thresholding the average optical flow of consecutive frames. Godard~\etal~\cite{godard2019digging} propose an auto-masking strategy to automatically mask out static pixels when calculating the photometric loss.

A second degenerate case is when the camera undergoes {\it purely rotational} motion. This degeneracy is well-known in traditional computer vision geometry~\cite{hartley2003multiple}, but has not been considered in self-supervised depth estimation. Under pure rotation, motion recovery using the fundamental matrix suffers from ambiguity, so homography-based methods are preferred~\cite{hartley2003multiple}. In the context of the photometric loss, if the camera motion is pure rotation, \ie, the translation ${\bf t} = 0$, the view synthesis (or reprojection) step 
does not depend on depth anymore (\ie, depth cancels out after applying the projection function). This is no surprise as their 2D correspondences are directly related by a homography matrix. So in this case, as long as the camera motion is accurately given, any arbitrary depth can minimize the photometric loss, which is undesirable when we train or finetune the depth network (as depth will be arbitrarily wrong).

To circumvent the degenerate cases described above, we propose a simple yet effective keyframe mechanism to facilitate online depth refinement without deterioration. After we receive camera poses from RAFT-SLAM, we can simply select keyframes for depth refinement according to the magnitude of camera {\it translations}. Only if the norm of the camera translation is over a certain threshold (see Sec.~\ref{sec:implementation}), we set its corresponding frame as a keyframe, \ie, the candidate for applying self-supervised losses. This ensures that we have enough baselines for the photometric loss to be effective.

\begin{algorithm}[!t]
  \small
  \caption{{\it GeoRefine}: self-supervised online depth refinement for geometrically consistent dense mapping.}
  \label{alg:georefine}
  \begin{algorithmic}[1]
     \State {\bf Pretrain} {the depth model. \Comment{supervised or self-supervised}}
     \State {Run RAFT-SLAM}. \Comment{on separate threads}
     \State {\bf Data preparation:} buffer time-synchronized keyframe data into a fixed-sized queue $\mathcal{Q}^*$; (optionally) form another data queue $\mathcal{Q}$ for per-frame data.
     \While{True}
     \State Check stop condition. \Comment{stop-signal from SLAM}
     \State Check SLAM failure signal. \Comment{clear data queue if received}
     \For{$k \gets 1$ to $K^*$} \Comment{{\it keyframe refinement}}
     \State Load data in $\mathcal{Q}^*$ to GPU, \Comment{batch size as 1}
     \State Compute losses as in Eq.~\eqref{loss:all},
     \State Update depth model via one gradient descent step. \Comment{ADAM optimizer}
     \EndFor
     
     \State Run inference and save refined depth for current keyframe.

     \For{$k \gets 1$ to $K$} \Comment{{\it Per-frame refinement}}
     \State Check camera translation from last frame, \Comment{skip if too small}
     \State Load data in $\mathcal{Q}^*$ and $\mathcal{Q}$ to GPU, \Comment{batch size as 1}
     \State Compute losses as in Eq.~\eqref{loss:all},
     \State Update depth model via one gradient descent step. \Comment{ADAM optimizer}
     \EndFor
     
     \State Run inference and save refined depth for current frame.
     
     \EndWhile
     \State{Run global mapping.} \Comment{TSDF or bundle fusion}
     \State{{\bf Output:} refined depth maps and global TSDF meshes.}
  \end{algorithmic}
\end{algorithm}

\subsubsection{Overall Refinement Strategy}\hfill\vspace{0.2cm}\\
Our overall refinement loss writes as 
\begin{equation}
\label{loss:all}
    L = L_p + \lambda_s L_s + \lambda_m L_m + \lambda_c L_c\;,
\end{equation}
where $\lambda_s, \lambda_m, \lambda_c$ are the weights balancing the contribution of each loss term.

GeoRefine aims to refine any pretrained depth models to achieve geometrically-consistent depth prediction for each frame of an image sequence. As RAFT-SLAM runs on separate threads, we buffer the keyframe data, including images, map points, and camera poses, into a time-synchronized data queue of a fixed size. If depth refinement is demanded for every frame, we additionally maintain a small data queue for per-frame data and construct the 5-frame snippet by taking 3 recent keyframes and 2 current consecutive frames. We conduct online refinement for the current keyframe (or frame) by minimizing the loss term in Eq.~\eqref{loss:all} and performing gradient descent for $K^*$ (or $K$) steps. After depth refinement steps, we run depth inference using the refined depth model and save the depth map for the current keyframe (or frame). Global maps can be finally reconstructed by performing TSDF or bundle fusion~\cite{niessner2013real,dai2017bundlefusion}. The whole {\it GeoRefine} algorithm is summarized in Alg.~\ref{alg:georefine}.

\section{Experiments}
\label{sec:exps}

We mainly conduct experiments on three public datasets: EuRoC~\cite{Burri25012016}, TUM-RGBD~\cite{sturm2012benchmark}, and ScanNet~\cite{dai2017scannet} datasets. Below, we first discuss the implementation details of our method. We then perform ablation studies to verify the effectiveness of each novel component in {\it GeoRefine}. Finally, we present quantitative and qualitative results on three datasets. For quantitative depth evaluation, we employ the standard error and accuracy metrics, including the Mean Absolute Error (MAE), Absolute Relative (\emph{Abs Rel}) error, \emph{RMSE}, $\delta < 1.25$ (namely \emph{$\delta_1$}), $\delta < 1.25^2$ (namely \emph{$\delta_2$}), and $\delta < 1.25^3$ (namely \emph{$\delta_3$}) as defined in~\cite{eigen2014depth}. 

\subsection{Implementation Details}
\label{sec:implementation}
Our GeoRefine includes a RAFT-SLAM module and an online depth refinement module. RAFT-SLAM is implemented based on ORB-SLAM3~\cite{campos2020orb} (other SLAM systems are also applicable) which support monocular, visual-inertial, and RGBD modes. In our experiments, we test the three modes and show that GeoRefine achieves consistent improvements over pretrained models. The pose data queue is maintained and updated in the SLAM side, where a frame pose is stored relative to its reference keyframe which is continuously optimized by BA and pose graph. 
The online learning module refines a pretrained depth model with customized data loader and training losses. In our experiments, we choose a supervised model, \ie, DPT~\cite{ranftl2021vision}, to showcase the effectiveness of our system. The initial DPT model is trained on a variety of public datasets and then finetuned on NYUv2~\cite{Silberman:ECCV12}. We utilize Robot Operating System (ROS)~\cite{ros} to exchange data between modules for cross-language compatibility.  We use ADAM~\cite{kingma2015adam} as the optimizer and set the learning rate to $1.0e^{-5}$. The weighting parameters $\lambda_s$, $\lambda_m$, and $\lambda_c$ are set to $1.0e^{-4}$, $5.0e^{-2}$, and $1.0e^{-1}$ respectively. 
We freeze its decoder layers of DPT for the sake of speed and stability. 
We filter map points with stringent criterion to ensure good supervision signal for online depth refinement. To this end, we discard map points observed in fewer than 5 keyframes or with reprojection errors greater than 1 pixel.
We maintain a keyframe data queue of length 11 and a per-frame data queue of length 2. The translation threshold for keyframe (or per-frame) refinement is set to 0.05 m (or 0.01 m). The number of refinement steps for keyframes (or per-frame) is set to 
3 (or 1). All system hyper-parameters are tuned on a validation sequence (EuRoC V2\_01).
Due to space limit, more system details of RAFT-SLAM, experimental results of our GeoRefine using the self-supervised Monodepth2~\cite{godard2019digging}, all experimental results on ScanNet, and other results are presented in the supplementary material.

\subsection{EuRoC Indoor MAV Dataset}
\label{sec:exp-euroc}

The EuRoC MAV dataset~\cite{Burri25012016} is an indoor dataset which provides stereo image sequences, IMU data and camera parameters. An MAV mounted with global shutter stereo cameras is used to capture the data in a large machine hall and a Vicon room. Five sequences are recorded in the machine hall and six are in the {Vicon} room. The ground-truth camera poses and depths are obtained with a {Vicon} device and a Leica MS50 laser scanner, so we use all Vicon sequences as the test set. 
We rectify the images with the provided intrinsics to remove image distortion. To generate ground-truth depths, we project the laser point cloud onto the image plane of the left camera using the code by~\cite{gordon2019depth}. The original images have a size of $480\times 754$ and are resized 
to $384\times 384$ for DPT.

\begin{table}[!t]
    \centering
    \caption{Quantitative depth evaluation on EuRoC under different SLAM modes.}
    \label{tab:euroc_all_eval}
    \resizebox{1.0\textwidth}{!}{
    \begin{tabular}{|l|c|c|c|c||c|c|c|c||c|c|c|c|}
    \hline
    \multirow{2}{*}{Method} & 
    \multicolumn{4}{c||}{Monocular} &  \multicolumn{4}{c||}{Visual-Inertial} &  \multicolumn{4}{c|}{pRGBD} \\
    \cline{2-13}
    ~ & MAE $\downarrow$ & AbsRel $\downarrow$ & RMSE $\downarrow$ & $\delta_1$ $\uparrow$ & MAE $\downarrow$ & AbsRel $\downarrow$ & RMSE $\downarrow$ & $\delta_1$ $\uparrow$ & MAE $\downarrow$ & AbsRel $\downarrow$ & RMSE $\downarrow$ & $\delta_1$ $\uparrow$ \\
    \hline
    \multicolumn{13}{c}{V1\_01} \\
    \hline
    DPT~\cite{ranftl2021vision} &   0.387  &   0.140  &  0.484   &    0.832  &   0.501  &   0.174  &  0.598   &  0.709 & 0.387  &   0.140  &  0.484   &    0.832 \\
    CodeMapping~\cite{matsuki2021codemapping} &   -  &   -  &  -   &    -  &  0.192   &  -    &   0.381  &   - & - & - & - & - \\
    Ours-DPT &   {\bf 0.153}  &   {\bf 0.050}  &   {\bf 0.241}  &  {\bf 0.980}  &  {\bf 0.147}  &   {\bf 0.048}  &  {\bf 0.241}   &  {\bf 0.980} & {\bf 0.151} & {\bf 0.049} & {\bf 0.239} & {\bf 0.982} \\
    \hline
    \multicolumn{13}{c}{V1\_02} \\
    \hline
    DPT~\cite{ranftl2021vision} &   0.320  &   0.119    &   0.412  &  0.882  &  0.496  &   0.182    &  0.586  &  0.712 & 0.320  &   0.119    &   0.412  &  0.882 \\
    CodeMapping~\cite{matsuki2021codemapping} &   -  &   -  &  -   &    -  &  0.259   &   -   &   0.369  &   -  & - & - & - & - \\
    Ours-DPT &   {\bf 0.171}  &  {\bf 0.058}   &   {\bf 0.255}  &  {\bf 0.967}  &   {\bf 0.166}  &   {\bf 0.058}   &  {\bf 0.251}  &  {\bf 0.972} & {\bf 0.160} & {\bf 0.056} & {\bf 0.240} & {\bf 0.973} \\
    \hline
    \multicolumn{13}{c}{V1\_03} \\
    \hline
    Monodepth2 \cite{godard2019digging} &   0.305  &   0.111    &   0.413  &   0.886  &   0.360  &   0.132  &   0.464  &  0.815 & 0.305  &   0.111    &   0.413  &   0.886 \\
    DPT~\cite{ranftl2021vision} &   0.305  &   0.112    &    0.396  &  0.890  &  0.499  &   0.185   &   0.581 & 0.700 & 0.305  &   0.112    &    0.396  &  0.890 \\
    CodeMapping~\cite{matsuki2021codemapping} &   -  &   -  &  -   &    -  &  0.283   &   -   &   0.407  &   - & - & - & - & - \\
    Ours-DPT &   {\bf 0.202}  &   {\bf 0.074}    &    {\bf 0.297}   &   {\bf 0.949}  &  {\bf 0.188}  &   {\bf 0.067}    &   {\bf 0.278}  & {\bf 0.956} & {\bf 0.190} & {\bf 0.068} & {\bf 0.286} & {\bf 0.949} \\
    \hline
    \multicolumn{13}{c}{V2\_01} \\
    \hline
    Monodepth2 \cite{godard2019digging} &   0.423  &   0.153    &    0.581  &   0.800  &  0.490  &   0.181  &   0.648   &   0.730 & 0.423  &   0.153    &    0.581  &   0.800 \\
    DPT~\cite{ranftl2021vision} &   0.325  &   0.128    &   0.436  &   0.854  &   0.482  &   0.205    &   0.571  &  0.703 & 0.325  &   0.128    &   0.436  &   0.854 \\
    CodeMapping~\cite{matsuki2021codemapping} &   -  &   -  &  -   &    -  &  0.290   &   -   &   0.428  &   - & - & - & - & - \\
    MonoIndoor~\cite{ji2021monoindoor} & - &  0.125 & 0.466 & 0.840 & - & - & - & - & - & - & - & - \\
    Ours-DPT &   {\bf 0.170}  &  {\bf 0.054}  &  {\bf 0.258}  &  {\bf 0.973}  &  {\bf 0.162}  &   {\bf 0.052}   &  {\bf 0.258}  &   {\bf 0.970} & {\bf 0.181} & {\bf 0.057} & {\bf 0.0269} & {\bf 0.970} \\
    \hline
    \multicolumn{13}{c}{V2\_02} \\
    \hline
    Monodepth2 \cite{godard2019digging} &   0.597  &   0.191    &   0.803  &   0.723  &    0.769  &   0.233  &   0.963 &   0.562 & 0.597  &   0.191    &   0.803  &   0.723 \\
    DPT~\cite{ranftl2021vision} &   0.404  &   0.134   &   0.540  & 0.838  &   0.601  &   0.191   &    0.727  &  0.699 & 0.404  &   0.134   &   0.540  & 0.838 \\
    CodeMapping~\cite{matsuki2021codemapping} &   -  &   -  &  -   &    -  &  0.415   &   -   &   0.655  &   - & - & - & - & - \\
    Ours-DPT &   {\bf 0.177}  &   {\bf 0.053}   &   {\bf 0.208}  &  {\bf 0.976}  &  {\bf 0.193}  &   {\bf 0.063}   &   {\bf 0.312}  & {\bf 0.966}  & {\bf 0.167} & {\bf 0.053} & {\bf 0.267} & {\bf 0.976} \\
    \hline
    \multicolumn{13}{c}{V2\_03} \\
    \hline
    Monodepth2 \cite{godard2019digging} &   0.601  &   0.211    &   0.784  &  0.673  &   0.764  &   0.258   &   0.912   &   0.498 & 0.601  &   0.211    &   0.784  &  0.673\\
    DPT~\cite{ranftl2021vision} &   0.283  &   0.099   &   0.366  &  0.905  &   0.480  &   0.154   &   0.564  & 0.746 & 0.283  &   0.099   &   0.366  &  0.905 \\
    CodeMapping~\cite{matsuki2021codemapping} &   -  &   -  &  -   &    -  &  0.686   &   -   &   0.952  &   -  & - & - & - & - \\
    Ours-DPT &   {\bf 0.163}  &   {\bf 0.053} & {\bf 0.231}   &  {\bf 0.970}  &  {\bf 0.159}  &   {\bf 0.055}   &   {\bf 0.220}  & {\bf 0.973} & {\bf 0.152} & {\bf 0.051} & {\bf 0.214} & {\bf 0.975} \\
    \hline
    \end{tabular}
    }
    \vspace{-0.4cm}
\end{table}

\noindent{\bf Quantitative Depth Results in the Monocular Mode.}
We conduct quantitative evaluation by running GeoRefine under {\it monocular} RAFT-SLAM on the EuRoC Vicon sequences, and present the depth evaluation results in the left columns of Tab.~\ref{tab:euroc_all_eval}. Following~\cite{godard2019digging}, we perform per-frame scale alignment between the depth prediction and the groundtruth.
From Tab.~\ref{tab:euroc_all_eval}, we can observe consistent and significant improvements by our method over the baseline model on all test sequences. In particular, on V1\_01, ``Ours-DPT'' reduces \emph{Abs Rel} from 14.0\% (by DPT) to 5.0\%, achieving over two-times reduction in depth errors. 

\noindent{\bf Quantitative Depth Results in the Visual-Inertial Mode.}
When IMU data are available, we can also run GeoRefine under {\it visual-inertial} (VI) RAFT-SLAM to get camera poses and map points directly in metric scale. Note that, in the visual-inertial mode, no scale alignment is needed. We present the quantitative depth results in Tab.~\ref{tab:euroc_all_eval}, from which we can see that our system under the VI mode performs on par with the monocular mode even without scale alignment. Compared to a similar dense mapping, \ie, CodeMapping~\cite{matsuki2021codemapping}, our GeoRefine is significantly more accurate with similar runtime (\ie, around 1 sec. per keyframe; see the supplementary), demonstrating the superiority of our system design.

\noindent{\bf Quantitative Depth Results in the pRGBD Mode.} We present the quantitative depth evaluation under the {\it pRGBD} mode in the right columns of Tab.~\ref{tab:euroc_all_eval}. We can see that the pRGBD mode performs slightly better than the other two modes in terms of depth results. This may be attributed to the fact that under this mode, the SLAM and depth refinement modules form a loosely-coupled loop so that each module benefits from the other.

\noindent{\bf Qualitative Depth Results.}
\begin{figure*}[!t]
\begin{center}
\includegraphics[width=1.0\textwidth]{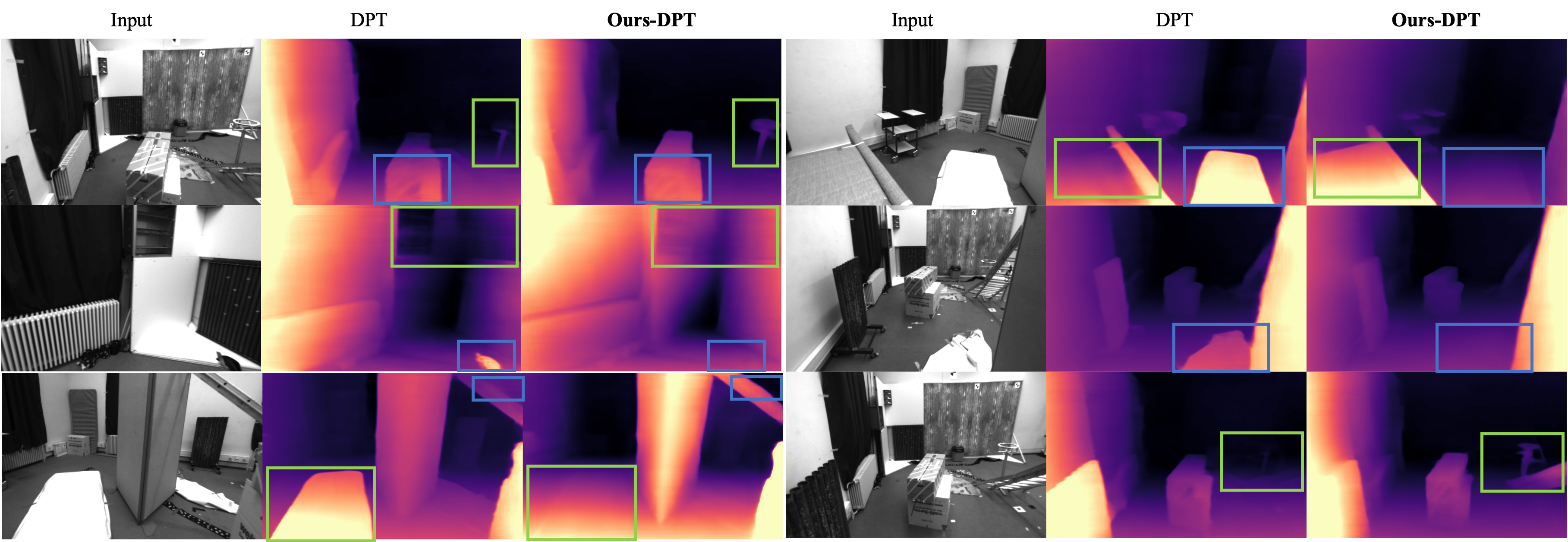}
\end{center}
\vspace{-0.6cm}
\caption{Visual comparison of depth maps by the pretrained DPT and our system. 
Regions with salient improvements are highlighted with green/blue boxes.}
\label{fig:vis_euroc}
\vspace{-0.1cm}
\end{figure*}
\begin{figure}[!t]
\begin{center}
\includegraphics[width=0.9\textwidth]{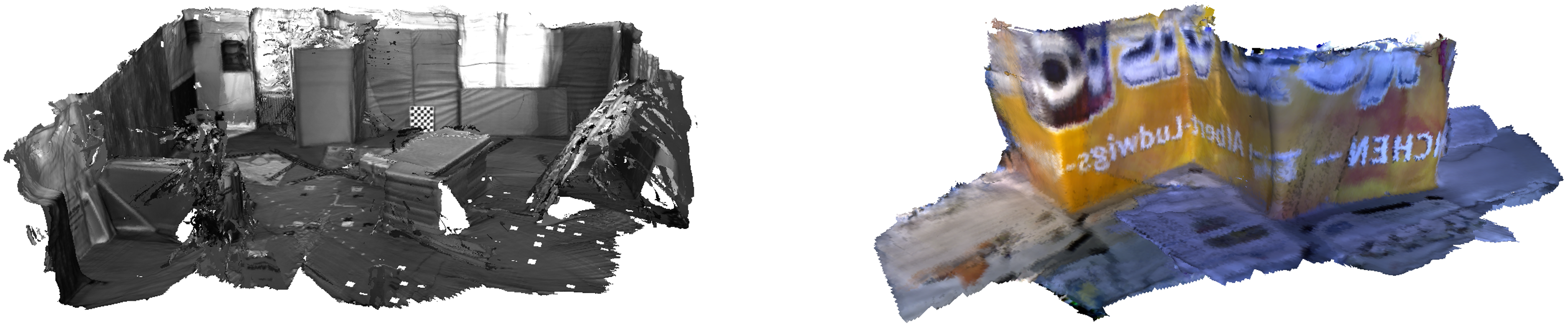}
\end{center}
\vspace{-0.7cm}
\caption{Global reconstruction on EuRoC (left) and TUM-RGBD (right) using the refined depth maps by GeoRefine.}
\label{fig:euroc_tum_fusion}
\vspace{-0.6cm}
\end{figure}
\begin{table*}[!t]
\caption{Monocular SLAM results on EuRoC (RMSE ATE in meters).}
\vspace{-0.3cm}
\begin{center}
\scalebox{0.60}{
\begin{tabular}{|p{3.2cm}|C{1.0cm} C{1.0cm} C{1.0cm} C{1.0cm} C{1.0cm} C{1.0cm} C{1.0cm} C{1.0cm} C{1.0cm} C{1.0cm} C{1.0cm}|C{1.0cm}|}
\hline
Method & MH\_01 & MH\_02 & MH\_03 & MH\_04 & MH\_05 & V1\_01 & V1\_02 & V1\_03 & V2\_01 & V2\_02 & V2\_03 & Mean \\
\hline
\hline
DeepFactor~\cite{czarnowski2020deepfactors} & 1.587 & 1.479 & 3.139 & 5.331 & 4.002 & 1.520 & 0.679 & 0.900 & 0.876 & 1.905 & 1.021 & 2.040 \\
DeepV2D~\cite{teed2018deepv2d} & 0.739 & 1.144 & 0.752 & 1.492 & 1.567 & 0.981 & 0.801 & 1.570 & 0.290 & 2.202 & 2.743 & 1.298 \\
D3VO~\cite{yang2020d3vo} & - & - & 0.080 & - & 0.090 & - & - & 0.110 & - & 0.050 & 0.019 & - \\
DROID-SLAM~\cite{teed2021droid} & 0.013 & {\bf 0.014} & {\bf 0.022} & {\bf 0.043} & 0.043 & 0.037 & 0.012 & {\bf 0.020} & {\bf 0.017} & 0.013 & {\bf 0.014} & {\bf 0.022}  \\
\hline
\hline
ORB-SLAM~\cite{mur2015orb} & 0.071 & 0.067 & 0.071 & 0.082 & 0.060 & {\bf 0.015} & 0.020 & x & 0.021 & 0.018 & x & - \\
DSO~\cite{engel2017direct} & 0.046 & 0.046 & 0.172 & 3.810 & 0.110 & 0.089 & 0.107 & 0.903 & 0.044 & 0.132 & 1.152 & 0.601 \\
ORB-SLAM3~\cite{campos2020orb} & 0.016 & 0.027 & 0.028 & 0.138 & 0.072 & 0.033 & 0.015 & 0.033 & 0.023 & 0.029 & x & -  \\
\hline
\hline
RAFT-SLAM (Ours) & {\bf 0.012} & 0.018 & 0.023 & 0.045 & {\bf 0.041} & 0.032 & {\bf 0.010} & 0.022 & 0.019 & {\bf 0.011} & 0.025 & 0.023 \\
\hline
\end{tabular}
}
\vspace{-0.8cm}
\end{center}
\label{tab:euroc_odometry}
\end{table*}
We show some visual comparisons in Fig.~\ref{fig:vis_euroc}, from which we can clearly observe the qualitative improvements brought by our online depth refinement method. 
In particular, our system can correct the inaccurate geometry that is commonly present in the pretrained model. For example, in the first row of Fig.~\ref{fig:vis_euroc}, a piece of thin paper lying on the floor is predicted to have much higher depth values than its neighboring floor pixels by the pretrained models (DPT); our GeoRefine is able to rectify its depth to be consistent with the floor. A global map of the EuRoC {VICON} room is shown in Fig.~\ref{fig:teaser} and~\ref{fig:euroc_tum_fusion}, where we can reach geometrically consistent reconstruction.

\noindent{\bf Odometry Results.}
Tab.~\ref{tab:euroc_odometry} shows the odometry comparisons of our proposed RAFT-SLAM with current state-of-the-art methods on the EuRoC dataset in the monocular mode. For fairness, we adopt the same parameter settings with ORB-SLAM3~\cite{campos2020orb} in all our experiments. Note that, although our system is not elaborately designed for SLAM, it achieves comparable results with DROID-SLAM~\cite{teed2021droid} and significantly outperforms other monocular baselines both in terms of accuracy and robustness.

\noindent{\bf Ablation Study.}
Without loss of generality, we perform an ablation study on Seq. V2\_03 to gauge the contribution of each component to our method under both monocular and pRGBD modes. Specifically, we first construct a base system by running a vanilla online refinement algorithm with the photometric loss as in Eq.~\eqref{loss:photo}, the depth smoothness loss as in Eq.~\eqref{loss:smooth}, and the map-point loss as in Eq.~\eqref{loss:mappoint}. Note that the photometric loss uses camera poses from RAFT-SLAM instead of a pose network. Under the monocular mode, we denote this base model as ``Our BaseSystem''. We then gradually add new components to this base model, including the RAFT-flow in SLAM front-end (``+RAFT-flow''), the scale alignment strategy in RAFT-SLAM (``+Scale Alignment''), and the occlusion-aware depth consistency loss (``+Depth Consistency''). Under the pRGBD mode, ``Our BaseSystem'' takes the pretrained depth as input without using our proposed changes, and this base system uses the depth consistency loss. We then gradually add new components to the base system, \ie, using refined depth from the online depth refinement module (``+Refined Depth''), using the RAFT-flow in SLAM front-end (``+RAFT-flow''), and removing the reprojection error term in bundle adjustment (``+Remove BA Term'').

\begin{table}[!t]
    \centering
    \caption{Ablation study on EuRoC Sequence V2\_03. Each component in our method improves the depth results.} 
    \label{tab:euroc_ablation}
    \begin{minipage}{0.495\linewidth}
    \resizebox{1.0\textwidth}{!}{
    {
    \begin{tabular}{|l|c|c|c|c|c|c|}
    \hline
    \multirow{2}{*}{Method} & 
    \multicolumn{6}{c|}{Monocular}  \\
    \cline{2-7} 
    ~  & MAE $\downarrow$ & Abs Rel $\downarrow$ & RMSE $\downarrow$  & $\delta_1$ $\uparrow$ & $\delta_2$ $\uparrow$ & $\delta_3$ $\uparrow$ \\ 
    \hline
    DPT~\cite{ranftl2021vision} &    0.283  &   0.099   &   0.366  &  0.905  &   0.979  &   0.994 \\ 
   Our BaseSystem & 0.269  &   0.090   &   0.347   &   0.905  &   0.983  &   0.997 \\ 
    + RAFT-flow & 0.248  &   0.083   &   0.331  &   0.915  &   0.985  &   0.997 \\ 
    + Scale Alignment  & 0.199  &   0.064   &   0.274   &   0.952  &   0.991  &   0.998 \\ 
    + Depth Consistency &   {\bf 0.163}  &   {\bf 0.053}   &   {\bf 0.231}    &   {\bf 0.970}  &   {\bf 0.995}  &   {\bf 0.999} \\ 
    \hline
    
    \end{tabular}
    }}
    \end{minipage}
    \hspace{-0.2cm}
    \begin{minipage}{0.495\linewidth}
    \resizebox{1.0\textwidth}{!}{
    {
    \begin{tabular}{|l|c|c|c|c|c|c|}
    \hline
    \multirow{2}{*}{Method} & 
    \multicolumn{6}{c|}{pRGBD}  \\
    \cline{2-7} 
    ~  & MAE $\downarrow$ & Abs Rel $\downarrow$ & RMSE $\downarrow$  & $\delta_1$ $\uparrow$ & $\delta_2$ $\uparrow$ & $\delta_3$ $\uparrow$ \\
    \hline
    DPT~\cite{ranftl2021vision} &    0.283  &   0.099   &   0.366  &  0.905  &   0.979  &   0.994 \\
    Our BaseSystem &   0.216  &   0.076   &   0.288   &   0.933  &   0.989  &   0.998  \\
    + Refined Depth & 0.199  &   0.065   &   0.268    &   0.958  &   0.995  &   {\bf 0.999} \\
    + RAFT-flow  &   0.171  &   0.056    &   0.237    &   0.972  &   0.995  &   0.998 \\
    + Remove BA Term \quad & {\bf 0.152}  &  {\bf 0.051}   &   {\bf 0.214}   & {\bf 0.975}  &   {\bf 0.997}  &   {\bf 0.999} \\
    \hline
    \end{tabular}
    }}
    \end{minipage}
    \vspace{-0.5cm}
\end{table}

We show a complete set of ablation results in Tab.~\ref{tab:euroc_ablation}. Under the monocular mode, ``Our BaseSystem'' reduces the absolute relative depth error from 9.9\% (by the pretrained DPT model) to 9.0\%, which verifies the effectiveness of the basic self-supervised refinement method. However, the improvement brought by our base model is not significant and the SLAM module fails. Using RAFT-flow in SLAM front-end makes SLAM more robust, generating more accurate pose estimation, which in turn improves the depth refinement module. Adding our scale self-alignment in RAFT-SLAM (``+Scale Alignment'') improves the depth quality significantly in all metrics, \eg, \emph{Abs Rel} decreases from $8.3\%$ to $6.4\%$ and $\delta_1$ increases from 91.5\% to 95.2\%.  Our occlusion-aware depth consistency loss (``Depth Consistency'') further achieves an improvement of 1.1\% in terms of \emph{Abs Rel} and 1.8\% in terms of $\delta_1$. From this ablation study, it is evident that each component of our method makes non-trivial contributions in improving depth results. We can draw a similar conclusion under the pRGBD mode.

\subsection{TUM-RGBD Dataset}

TUM-RGBD is a well-known dataset mainly for benchmarking performance of RGB-D SLAM or odometry~\cite{sturm2012benchmark}. This dataset was created using a Microsoft Kinect sensor and eight high-speed tracking cameras to capture monocular images, their corresponding depth images, and camera poses. This dataset is particularly difficult for monocular systems as it contains a large amount of motion blur and rolling-shutter distortion caused by fast camera motion. We take two monocular sequences from this dataset, \ie, ``freiburg3\_structure\_texture\_near'' and ``freiburg3\_structure\_texture\_far'', to test our system, as they satisfy our system's requirement of sufficient camera translations. 
The quantitative depth results are presented in Tab.~\ref{tab:tum_all_eval}. As before, under both SLAM modes, our GeoRefine improves upon the pretrained DPT model by a significant margin, achieving 2-4 times' reduction in terms of {\it Abs Rel}. A global reconstruction is visualized in Fig.~\ref{fig:euroc_tum_fusion}, where the scene geometry is faithfully recovered. See more in the suppl..

\begin{table}[!t]
    \centering
    \caption{Quantitative depth evaluation on TUM-RGBD.}
    \label{tab:tum_all_eval}
    \resizebox{0.8\textwidth}{!}{
    \begin{tabular}{|l|c|c|c|c|c|c||c|c|c|c|c|c|}
    \hline
    \multirow{2}{*}{Method} & 
    \multicolumn{6}{c||}{Monocular} &  \multicolumn{6}{c|}{pRGBD} \\
    \cline{2-13}
    ~ & MAE $\downarrow$ & AbsRel $\downarrow$ & RMSE $\downarrow$ & $\delta_1$ $\uparrow$ & $\delta_2$ $\uparrow$ & $\delta_3$ $\uparrow$ & MAE $\downarrow$ & AbsRel $\downarrow$ & RMSE $\downarrow$  & $\delta_1$ $\uparrow$ & $\delta_2$ $\uparrow$ & $\delta_3$ $\uparrow$ \\
    \hline
    \multicolumn{13}{c}{freiburg3\_structure\_texture\_near} \\
    \hline
    DPT~\cite{ranftl2021vision} &   0.280  &   0.140    &   0.529  &  0.794  &   0.924  &   0.968 & 0.280  &   0.140    &   0.529  &  0.794  &   0.924  &   0.968   \\
    Ours-DPT &   {\bf 0.138}  &   {\bf 0.057}  &   {\bf 0.314}  &  {\bf 0.943}  &  {\bf 0.977}  &  {\bf 0.990}  &{\bf 0.140}  &   {\bf 0.056}  &   {\bf 0.317}  &  {\bf 0.941}  &  {\bf 0.974}  &  {\bf 0.992} \\
    \hline
    \multicolumn{13}{c}{freiburg3\_structure\_texture\_far} \\
    \hline 
    DPT~\cite{ranftl2021vision} &   0.372  &   0.134  &  0.694   &  0.810  &   0.939  &   0.968  & 0.372  &   0.134  &  0.694   &  0.810  &   0.939  &   0.968  \\
    Ours-DPT &   {\bf 0.108}  &  {\bf 0.035}   &   {\bf 0.317}  &  {\bf 0.974}  &   {\bf 0.985}  &  {\bf 0.997}  & {\bf 0.105} & {\bf 0.036} & {\bf 0.290} & {\bf 0.975} & {\bf 0.985} & {\bf 0.996} \\
    \hline
    \end{tabular}
    }
    \vspace{-0.6cm}
\end{table}


\section{Conclusions}
\label{sec:cons}
In this paper, we have introduced {\it GeoRefine}, an online depth refinement system that combines geometry and deep learning. 
The core contribution of this work lies in the system design itself, where we show that accurate dense mapping from monocular sequences is possible via a robust hybrid SLAM, an online learning paradigm, and a careful consideration of degenerate cases. The self-supervised nature of the proposed system also suggests that it can be deployed in any unseen environments by virtue of its self-adaptation capability.
We have demonstrated the state-of-the-art performance on several challenging public datasets.

\noindent{\bf Limitations.} Our system does not have a robust mechanism to handle moving objects which are outliers both for SLAM and self-supervised losses. Hence, datasets with plenty of foreground moving objects such as KITTI \cite{Geiger2013IJRR} would not be the best test-bed for GeoRefine. 
Another limitation is that GeoRefine cannot deal with scenarios where camera translations are small over the entire sequence. This constraint is intrinsic to our system design, but it is worth exploring how to relax it while maintaining robustness.

\section*{Appendix}

\subsection{Implementation Details}

In this section, we describe the implementation details of RAFT-SLAM and present a simple mechanism to handle SLAM failures.

\subsubsection{RAFT-SLAM}

Our system utilizes ROS as the agent for cross-language communication. 
Consecutive frames are fed into the RAFT network~\cite{teed2020raft} to get pair-wise flow predictions, including both the forward and the backward flows. For all our experiments, we use the RAFT flow model that is pretrained on FlyingThings3D, \ie, raft-things.pth, downloaded from https://github.com/princeton-vl/RAFT. 
In the monocular mode, after the system successfully initializes, we continuously align the map points and camera poses to CNN depth for five steps to make their scales consistent to each other.

\subsubsection{SLAM Failures}

It is hard to ensure RAFT-SLAM never encounters failure cases. We observe that it fails occasionally on sequences with strong motion blur and significant rolling-shutter artifacts. In the event of SLAM failures, we want the depth model to be rarely disrupted and the system is supposed to continue to run after the SLAM module recovers. To this end, we employ a simple strategy, \ie, after the depth refinement module receives a signal of SLAM failure, the system clears the queues both for keyframe and per-frame data. In this case, the keyframe depth refinement process is paused, but the per-frame depth inference can still run if depth maps for all frames are demanded.

\begin{figure}[!t]
\begin{center}
\includegraphics[width=1.0\textwidth]{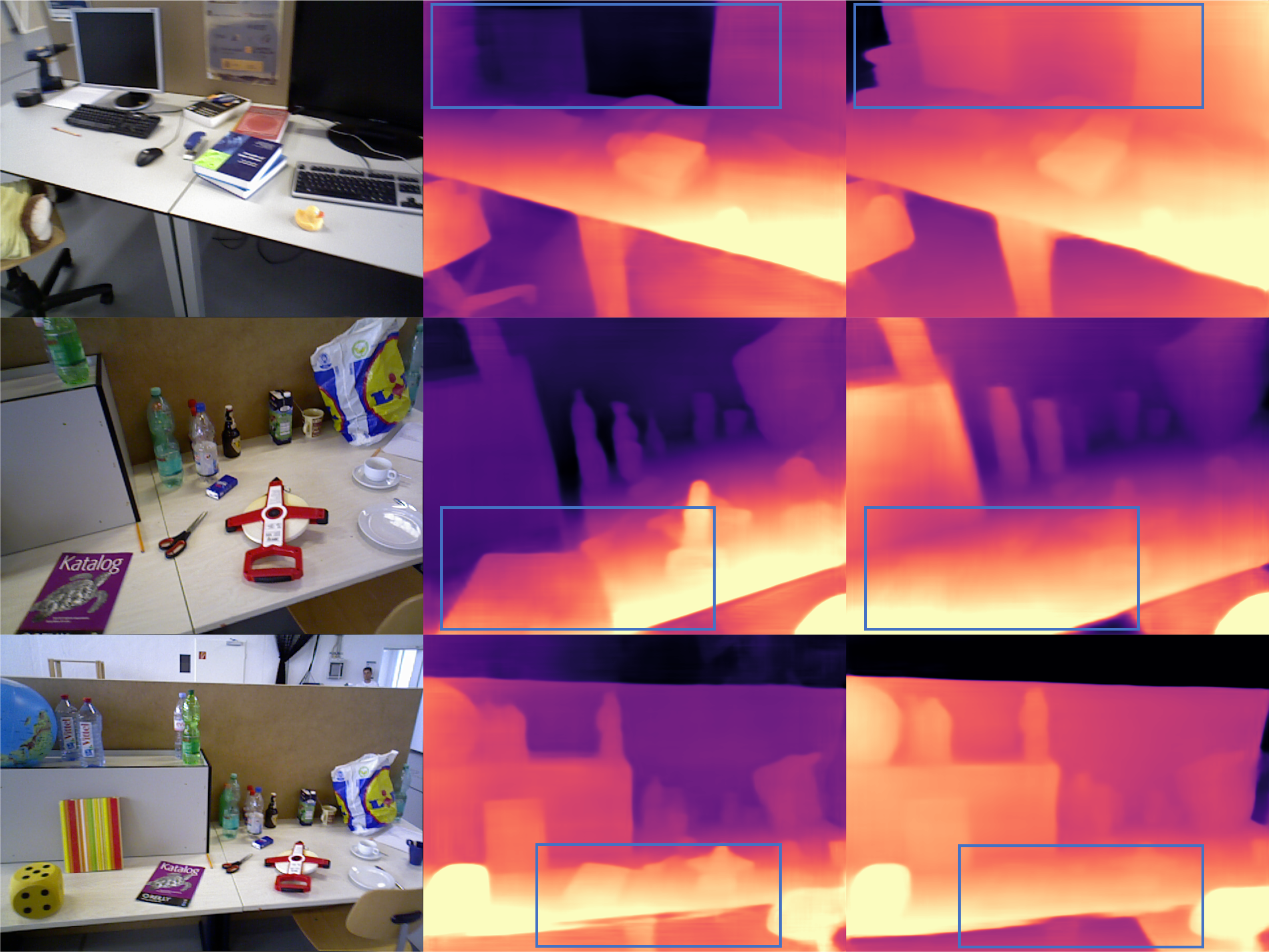}
\end{center}
\vspace{-0.6cm}
\caption{Qualitative results on TUM-RGBD. From left to right: input images, depth maps by DPT, depth maps by our GeoRefine. Our method is able to eliminate many artifacts and erroneous predictions compared to DPT.}
\label{fig:tum_qualitative}
\vspace{-0.4cm}
\end{figure}

\subsection{EuRoC}

In this section, we include additional depth and pose results on EuRoC. More qualitative results can be found in the attached videos.

\subsubsection{GeoRefine-MD2}
We present depth results of GeoRefine using a self-supervised model, \ie, Monodepth2~\cite{godard2019digging}, as the base model on EuRoC. We take monocular and stereo images from five sequences (MH\_01, MH\_02, MH\_04, V1\_01, and V1\_02) as the training set to train the base model Monodepth2. Since stereo images with a known baseline distance are used, the pretrained Monodepth2 is scale-aware. The quantitative depth results are shown in Tab.~\ref{tab:euroc_all_eval_md2}, from which we can see that our system, denoted as ``Ours-MD2'', improves over Monodepth2 by a significant margin in all three SLAM modes. 

\begin{table*}[!t]
\caption{pRGBD SLAM results on EuRoC (RMSE ATE in meters).}
\vspace{-0.3cm}
\begin{center}
\scalebox{0.75}{
\begin{tabular}{|l|C{1.0cm} | C{1.0cm} | C{1.0cm} | C{1.0cm} | C{1.0cm} | C{1.0cm} | C{1.0cm} | C{1.0cm} | C{1.0cm} | C{1.0cm} | C{1.0cm} || C{1.0cm}|}
\hline
Method & MH\_01 & MH\_02 & MH\_03 & MH\_04 & MH\_05 & V1\_01 & V1\_02 & V1\_03 & V2\_01 & V2\_02 & V2\_03 & Mean \\
\hline
\hline
ORB-SLAM3~\cite{campos2020orb} & {\bf 0.016} & 0.027 & {\bf 0.028} & 0.138 & 0.072 & {\bf 0.033} & {\bf 0.015} & 0.033 & 0.023 & 0.029 & x & -  \\
Ours-pRGBD & 0.025 & {\bf 0.023} & 0.031 & {\bf 0.064} & {\bf 0.060} & {\bf 0.033} & {\bf 0.015} & {\bf 0.023} & {\bf 0.022} & {\bf 0.016} & {\bf 0.034} & {\bf 0.031}\\
\hline
\end{tabular}
}
\vspace{-0.8cm}
\end{center}
\label{tab:euroc_odometry_prgbd}
\end{table*}
\subsubsection{Odometry and Ablation}
\begin{figure}[!t]
     \centering
     \begin{subfigure}[b]{0.33\textwidth}
         \centering
         \includegraphics[width=\textwidth]{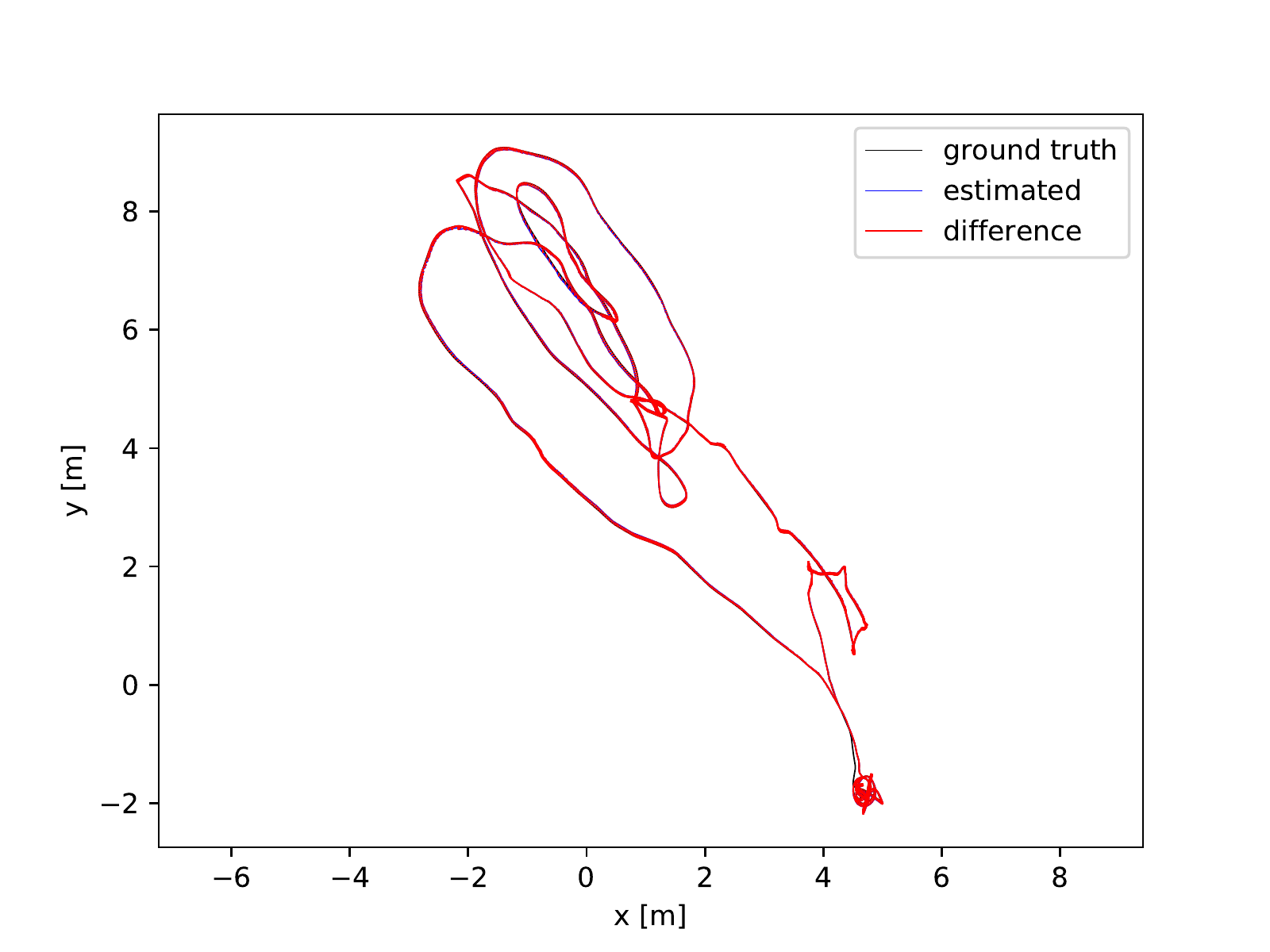}
         \vspace{-0.4cm}
         \caption{MH\_01}
         \label{fig:mh_01}
     \end{subfigure}
     \hspace{-0.4cm}
     \begin{subfigure}[b]{0.33\textwidth}
         \centering
         \includegraphics[width=\textwidth]{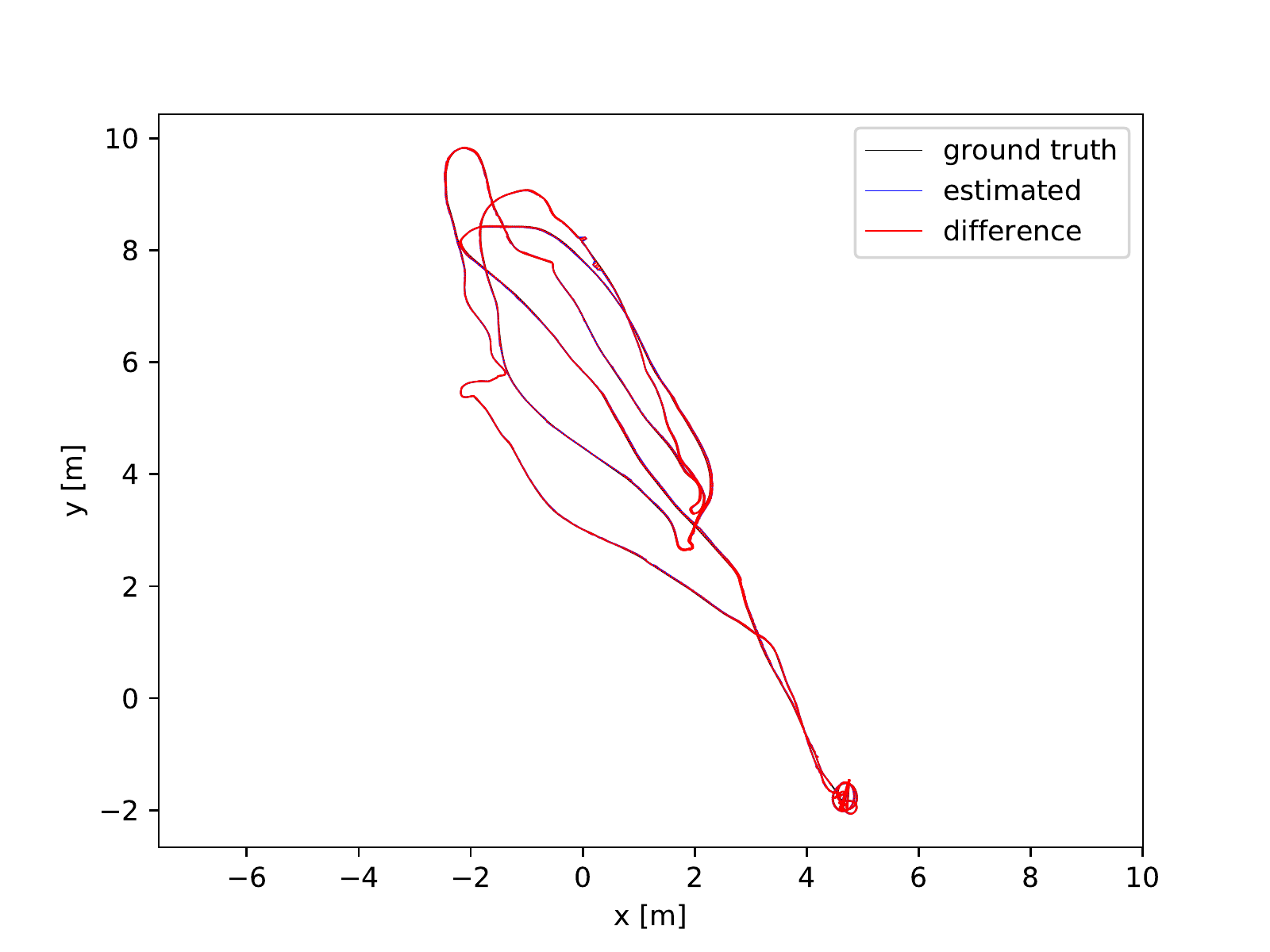}
         \vspace{-0.4cm}
         \caption{MH\_02}
         \label{fig:mh_02}
     \end{subfigure}
     \hspace{-0.4cm}
     \begin{subfigure}[b]{0.33\textwidth}
         \centering
         \includegraphics[width=\textwidth]{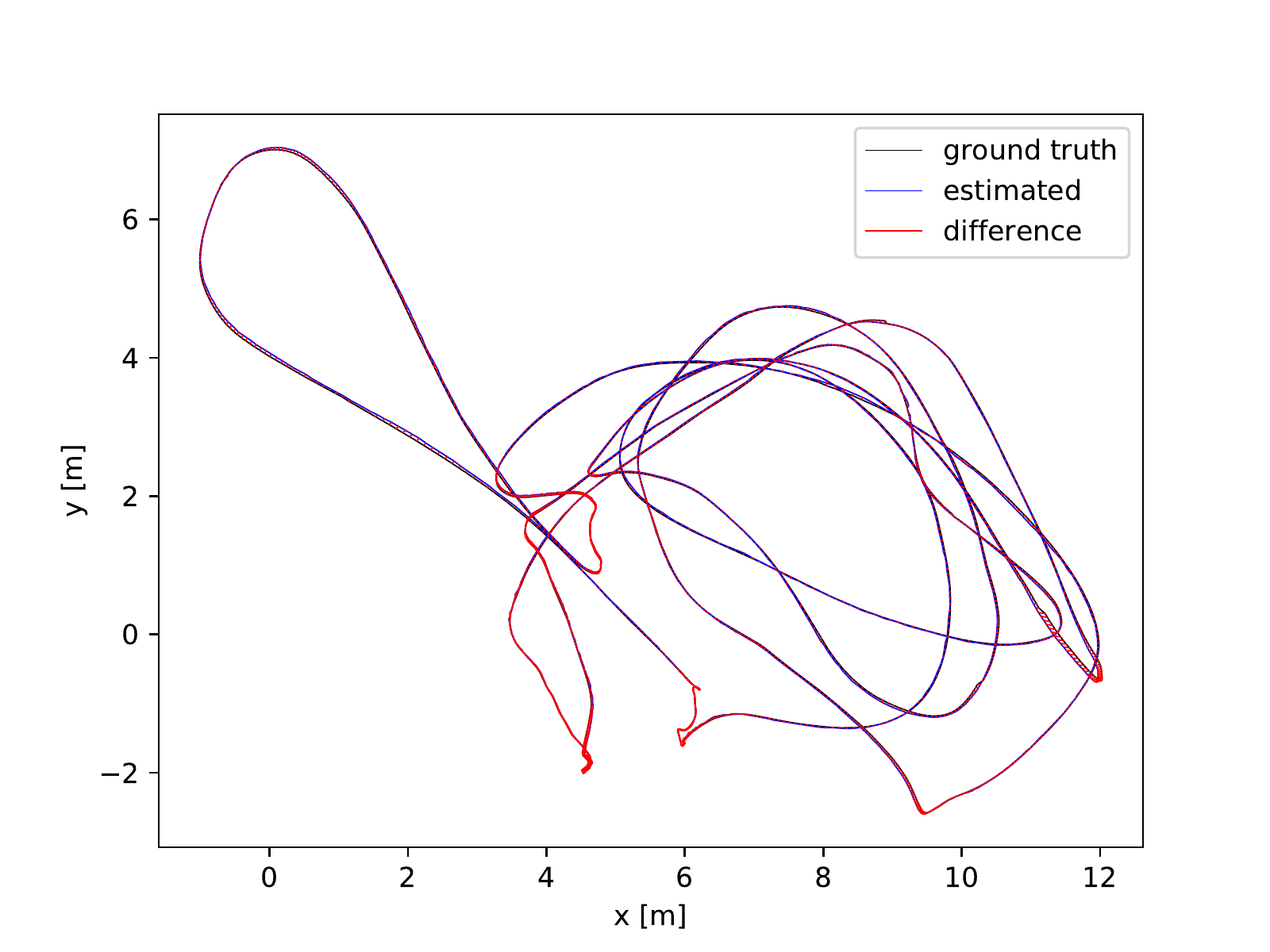}
         \vspace{-0.4cm}
         \caption{MH\_03}
         \label{fig:mh_03}
     \end{subfigure}
     \hspace{-0.4cm}
     \begin{subfigure}[b]{0.33\textwidth}
         \centering
         \includegraphics[width=\textwidth]{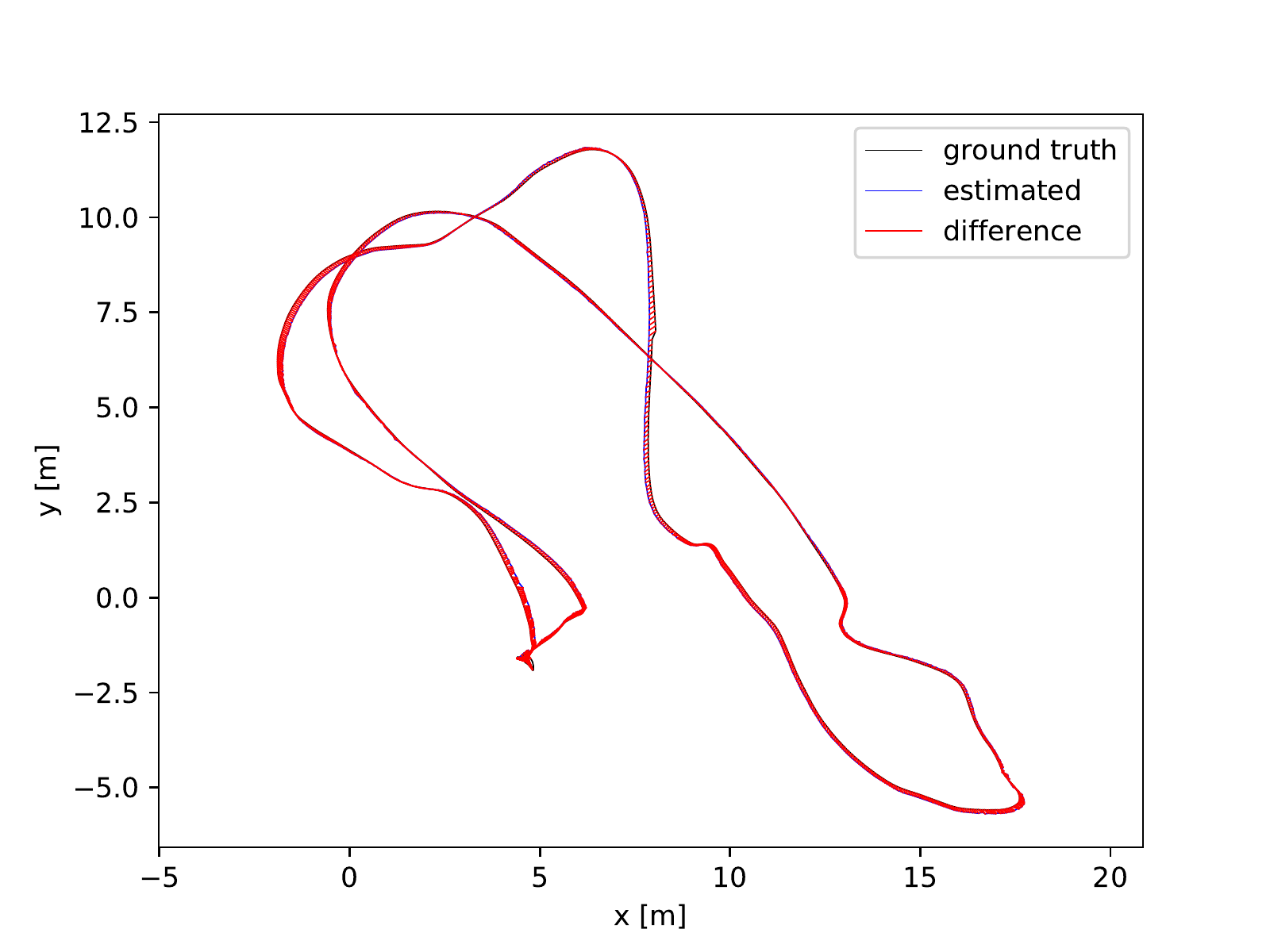}
         \vspace{-0.4cm}
         \caption{MH\_04}
         \label{fig:mh_04}
     \end{subfigure}
     \hspace{-0.4cm}
     \begin{subfigure}[b]{0.33\textwidth}
         \centering
         \includegraphics[width=\textwidth]{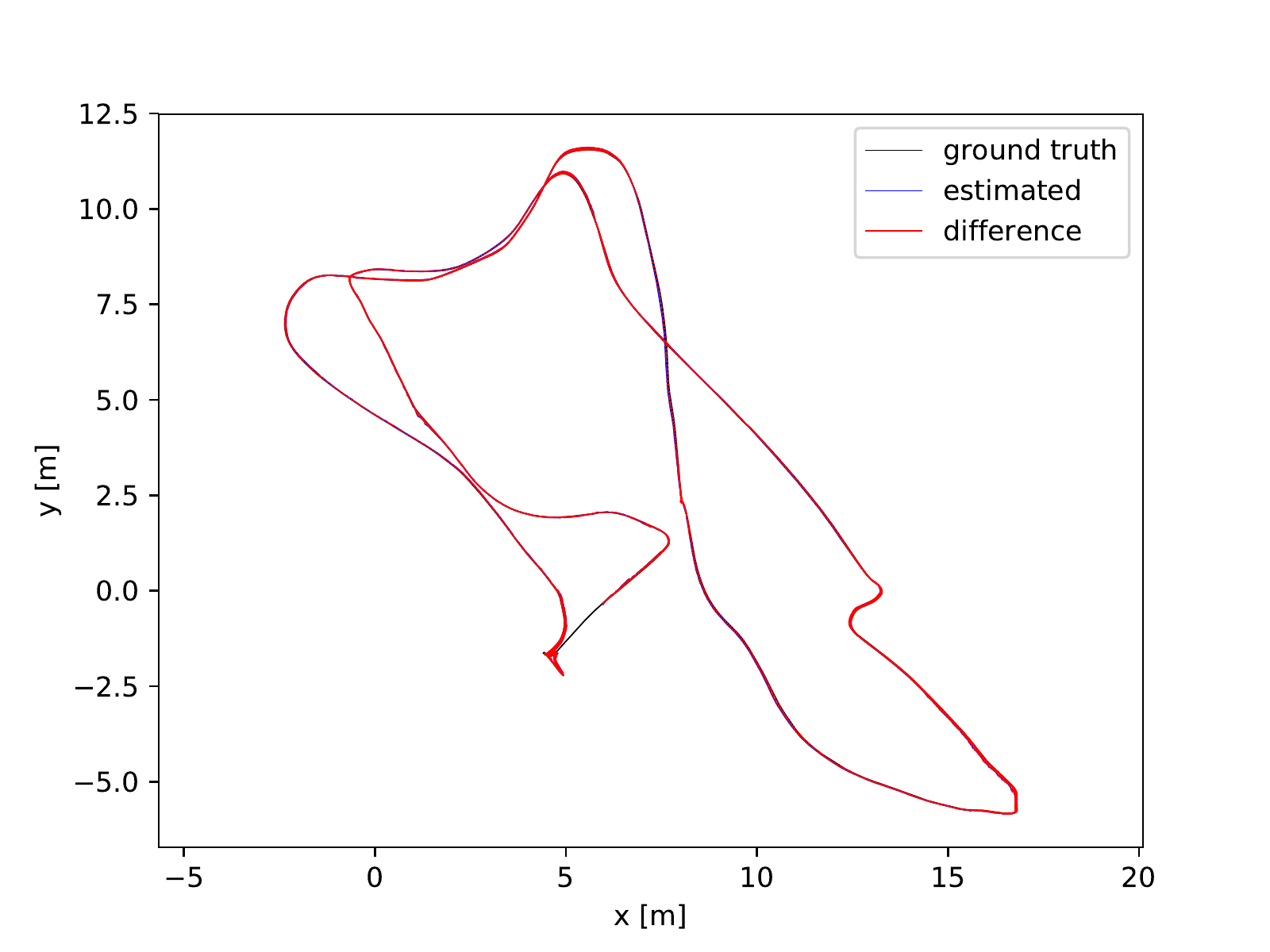}
         \vspace{-0.4cm}
         \caption{MH\_05}
         \label{fig:mh_05}
     \end{subfigure}
     \vspace{-0.2cm}
     \caption{Visualized trajectory results on EuRoC MH sequences. Best viewed on screen with zoom-in.}
     \vspace{-0.4cm}
     \label{fig:mh_traj}
\end{figure}

Tab.~\ref{tab:euroc_odometry_prgbd} and Tab.~\ref{tab:euroc_ablation_all} report the odometry results of our proposed RAFT-SLAM in the pRGBD mode and the corresponding ablation study. It's evident that our pRGBD RAFT-SLAM outperforms the baseline, \ie, ORB-SLAM3, both in terms of robustness and accuracy, and each proposed new component contributes to the improvement. Note that ``Our BaseSystem'' uses only the pretrained depth from DPT to form a pRGBD mode. Fig.~\ref{fig:mh_traj} shows the visualized trajectories on EuRoC MH sequences.

\begin{table}[!t]
    \centering
    \caption{Ablation study on EuRoC Sequence V2\_03 in pRGBD mode} 
    \label{tab:euroc_ablation_all}
    \resizebox{0.8\textwidth}{!}{
    {
    \begin{tabular}{|l|c|c|c|c|c|c||c|}
    \hline
    \multirow{2}{*}{Method} & 
    \multicolumn{6}{c||}{Depth} & \multicolumn{1}{c|}{Odometry} \\
    \cline{2-8} 
    ~  & MAE $\downarrow$ & Abs Rel $\downarrow$ & RMSE $\downarrow$  & $\delta_1$ $\uparrow$ & $\delta_2$ $\uparrow$ & $\delta_3$ $\uparrow$ & RMSE ATE $\downarrow$ \\
    \hline
    DPT~\cite{ranftl2021vision} &    0.283  &   0.099   &   0.366  &  0.905  &   0.979  &   0.994 & - \\
    Our BaseSystem &   0.216  &   0.076   &   0.288   &   0.933  &   0.989  &   0.998  & 0.176 \\
    + Refined Depth & 0.199  &   0.065   &   0.268    &   0.958  &   0.995  &   {\bf 0.999} & 0.133 \\
    + RAFT-flow  &   0.171  &   0.056    &   0.237    &   0.972  &   0.995  &   0.998 & 0.069\\
    + Remove BA Term \quad & {\bf 0.152}  &  {\bf 0.051}   &   {\bf 0.214}   & {\bf 0.975}  &   {\bf 0.997}  &   {\bf 0.999} & {\bf 0.034}\\
    \hline
    \end{tabular}
    }}
    \vspace{-0.4cm}
\end{table}




\begin{table}[!t]
    \centering
    \caption{Quantitative depth evaluation on EuRoC using Monodepth2.}
    \label{tab:euroc_all_eval_md2}
    \resizebox{1.0\textwidth}{!}{
    \begin{tabular}{|l|c|c|c|c||c|c|c|c||c|c|c|c|}
    \hline
    \multirow{2}{*}{Method} & 
    \multicolumn{4}{c||}{Monocular} &  \multicolumn{4}{c||}{Visual-Inertial} &  \multicolumn{4}{c|}{pRGBD} \\
    \cline{2-13}
    ~ & MAE $\downarrow$ & AbsRel $\downarrow$ & RMSE $\downarrow$ & $\delta_1$ $\uparrow$ & MAE $\downarrow$ & AbsRel $\downarrow$ & RMSE $\downarrow$ & $\delta_1$ $\uparrow$ & MAE $\downarrow$ & AbsRel $\downarrow$ & RMSE $\downarrow$ & $\delta_1$ $\uparrow$ \\
    \hline
    \multicolumn{13}{c}{V1\_03} \\
    \hline
    Monodepth2 \cite{godard2019digging} &   0.305  &   0.111    &   0.413  &   0.886  &   0.360  &   0.132  &   0.464  &  0.815 & 0.305  &   0.111    &   0.413  &   0.886 \\
    Ours-MD2 &   {\bf 0.184}  &   {\bf 0.066}   &    {\bf 0.272}  & {\bf 0.960}  &   {\bf 0.178}  &   {\bf 0.062}    &   {\bf 0.255}  &  {\bf 0.972} &  {\bf 0.178}  &   {\bf 0.059}  &   {\bf 0.251}   &   {\bf 0.966} \\
    \hline
    \multicolumn{13}{c}{V2\_01} \\
    \hline
    Monodepth2 \cite{godard2019digging} &   0.423  &   0.153    &    0.581  &   0.800  &  0.490  &   0.181  &   0.648   &   0.730 & 0.423  &   0.153    &    0.581  &   0.800 \\
    Ours-MD2 &   {\bf 0.202}  &   {\bf 0.063}    &   {\bf 0.306}  &  {\bf 0.960}  &   {\bf 0.169}  &   {\bf 0.059}   &  {\bf 0.265}  &  {\bf 0.968} & {\bf 0.191}  &   {\bf 0.060}  &   {\bf 0.295}   &   {\bf 0.958} \\
    \hline
    \multicolumn{13}{c}{V2\_02} \\
    \hline
    Monodepth2 \cite{godard2019digging} &   0.597  &   0.191    &   0.803  &   0.723  &    0.769  &   0.233  &   0.963 &   0.562 & 0.597  &   0.191    &   0.803  &   0.723 \\
    Ours-MD2 &   {\bf 0.218}  &   {\bf 0.065}    &   {\bf 0.350}  &  {\bf 0.955}  &  {\bf 0.193}  &   {\bf 0.060}    &   {\bf 0.320}  & {\bf 0.964} &  {\bf 0.199}  &   {\bf 0.059}   &   {\bf 0.327}   &   {\bf 0.962} \\
    \hline
    \multicolumn{13}{c}{V2\_03} \\
    \hline
    Monodepth2 \cite{godard2019digging} &   0.601  &   0.211    &   0.784  &  0.673  &   0.764  &   0.258   &   0.912   &   0.498 & 0.601  &   0.211    &   0.784  &  0.673\\
    Ours-MD2 &   {\bf 0.192}  &   {\bf 0.064}    &   {\bf 0.266}  &   {\bf 0.956}  &   {\bf 0.171}  &   {\bf 0.059}   &   {\bf 0.251}  & {\bf 0.968} & {\bf 0.207} & {\bf 0.069} & {\bf 0.297} & {\bf 0.951} \\
    \hline
    \end{tabular}
    }
\end{table}


\subsection{TUM-RGBD}

\begin{figure}[!t]
\begin{center}
\includegraphics[width=1.0\textwidth]{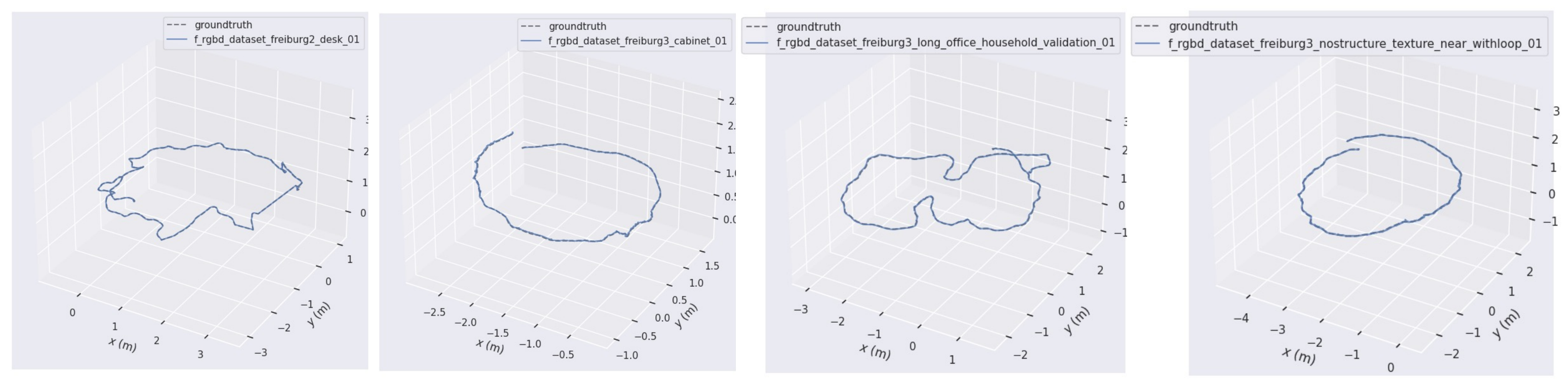}
\end{center}
\vspace{-0.6cm}
\caption{Qualitative pose results of our system under the pRGBD mode on TUM-RGBD. Best viewed on screen with zoom-in.}
\label{fig:tum_pose}
\vspace{-0.4cm}
\end{figure}
\begin{table}[!t]
\vspace{-0.3cm}
    {\caption{Odometry results on TUM-RGBD in terms of {\bf RPE [m/s]}. ``X'' means no pose output due to system failure and ``(X)'' means partial pose results.}\label{tab:odo-tum}}
    \centering
    \resizebox{1.0\textwidth}{!}{
    \begin{tabular}{|c|c|c|c|c|c|c|c|c||c|c|}
    \hline
    Method & f2/desk & f2/pio\_360 & 
    f2/pio\_slam & f3/cbnet &  f3/l\_o\_h\_val & f3/ns\_t\_nr\_lp & f3/str\_nt\_f & f3/str\_nt\_n & mean \\ 
    \hline
    ORB-SLAM3~\cite{campos2020orb} & 0.039 & 0.155(X)  & X  & 0.160(X) & 0.024 & 0.604 & X & X & - \\ 
    Li~\cite{li2021generalizing} & 0.158 & 0.201 & 0.176  & 0.213 & 0.133 & 0.159 & 0.104 & 0.207 & 0.169 \\ 
    Ours-Mono & \textbf{0.025} & \textbf{0.075} & 0.161  & 0.079 & \textbf{0.022} & {\bf 0.028} & 
    0.107 & 0.195(X) & 0.089 \\
    Ours-pGRBD & 0.033 & 0.092 & {\bf 0.133}  & \textbf{0.023} & 0.028 & 0.031 & {\bf 0.042} & {\bf 0.092} & {\bf 0.059} \\ 
    \hline
    \end{tabular}
    }
\vspace{-0.4cm}
\end{table}

\begin{figure*}[!t]
\begin{center}
\includegraphics[width=1.0\textwidth]{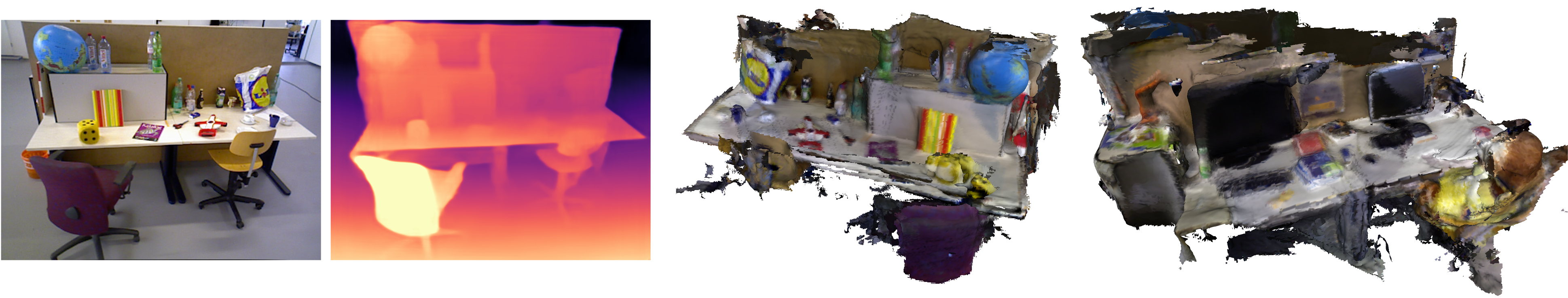}
\end{center}
\vspace{-0.5cm}
\caption{Global reconstruction on TUM-RGBD (freiburg3\_long\_office\_household) using the refined depth maps by GeoRefine.}
\label{fig:tum_fusion_suppl}
\vspace{-0.4cm}
\end{figure*}

We evaluate our GeoRefine on a few more sequences 
from the TUM-RGBD dataset. We adopt the same settings as in the main paper and use the DPT model~\cite{ranftl2021vision} pretrained on NYUv2 as our initial model. The quantitative depth results are shown in Tab.~\ref{tab:tum_all_eval_sup}, from which we can observe consistent and significant improvements by our GeoRefine over the pretrained model. Qualitative results can be found in Fig.~\ref{fig:tum_qualitative}, Fig.~\ref{fig:tum_fusion_suppl} and the attached video.

In addition, we compare with~\cite{li2021generalizing} and show odometry results in terms of relative pose error (RPE) on TUM-RGBD in Tab.~\ref{tab:odo-tum}. Compared to the baseline ORB-SLAM3~\cite{campos2020orb}, the improved odometry results by our system verify that {\it using RAFT makes the SLAM system more robust and accurate}. In particular, our method in both the monocular and pRGBD modes outperforms a recent deep odometry method~\cite{li2021generalizing} by a significant margin. 
See Fig.~\ref{fig:tum_pose} for qualitative pose results of our system under the pRGBD mode.


\begin{table}[!t]
    \centering
    \caption{Quantitative depth evaluation on additional TUM-RGBD sequences.}
    \label{tab:tum_all_eval_sup}
    \resizebox{1.0\textwidth}{!}{
    \begin{tabular}{|l|c|c|c|c|c|c||c|c|c|c|c|c|}
    \hline
    \multirow{2}{*}{Method} & 
    \multicolumn{6}{c||}{Monocular} &  \multicolumn{6}{c|}{pRGBD} \\
    \cline{2-13}
    ~ & MAE $\downarrow$ & AbsRel $\downarrow$ & RMSE $\downarrow$ & $\delta_1$ $\uparrow$ & $\delta_2$ $\uparrow$ & $\delta_3$ $\uparrow$ & MAE $\downarrow$ & AbsRel $\downarrow$ & RMSE $\downarrow$  & $\delta_1$ $\uparrow$ & $\delta_2$ $\uparrow$ & $\delta_3$ $\uparrow$ \\
    \hline
    \multicolumn{13}{c}{freiburg3\_long\_office\_household} \\
    \hline
    DPT~\cite{ranftl2021vision} &   0.366  &   0.129     &   0.762  &  0.833  &   0.926  &   0.955 & 0.366  &   0.129     &   0.762  &  0.833  &   0.926  &   0.955   \\
    Ours-DPT &   {\bf 0.175}  &   {\bf 0.078}  &   {\bf 0.349}  &  {\bf 0.926}  &  {\bf 0.973}  &  {\bf 0.993}  &{\bf 0.146}  &   {\bf 0.065}  &   {\bf 0.315}  &  {\bf 0.947}  &  {\bf 0.989}  &  {\bf 0.997} \\
    \hline
    \multicolumn{13}{c}{freiburg3\_long\_office\_household\_validation} \\
    \hline 
    DPT~\cite{ranftl2021vision} &   0.350  &   0.136  &  0.750   &  0.836  &   0.924  &   0.948  & 0.350  &   0.136  &  0.750   &  0.836  &   0.924  &   0.948  \\
    Ours-DPT &   {\bf 0.171}  &  {\bf 0.078}   &   {\bf 0.380}  &  {\bf 0.930}  &   {\bf 0.965}  &  {\bf 0.976}  & {\bf 0.151} & {\bf 0.071} & {\bf 0.341} & {\bf 0.941} & {\bf 0.977} & {\bf 0.993} \\
    \hline
    \multicolumn{13}{c}{freiburg3\_nostructure\_texture\_near\_withloop} \\
    \hline 
    DPT~\cite{ranftl2021vision} &   0.129  &   0.103   &   0.163   &   0.914  &   0.999  &   1.000  & 0.129  &   0.103   &   0.163   &   0.914  &   0.999  &   1.000  \\
    Ours-DPT &   {\bf 0.028}  &  {\bf 0.024}   &   {\bf 0.039}  &  {\bf 0.996}  &   {\bf 1.000}  &  {\bf 1.000}  & {\bf 0.028} & {\bf 0.024} & {\bf 0.039} & {\bf 1.000} & {\bf 1.000} & {\bf 1.000} \\
    \hline
    \end{tabular}
    }
\end{table}

\begin{figure*}[!t]
\begin{center}
\includegraphics[width=1.0\textwidth]{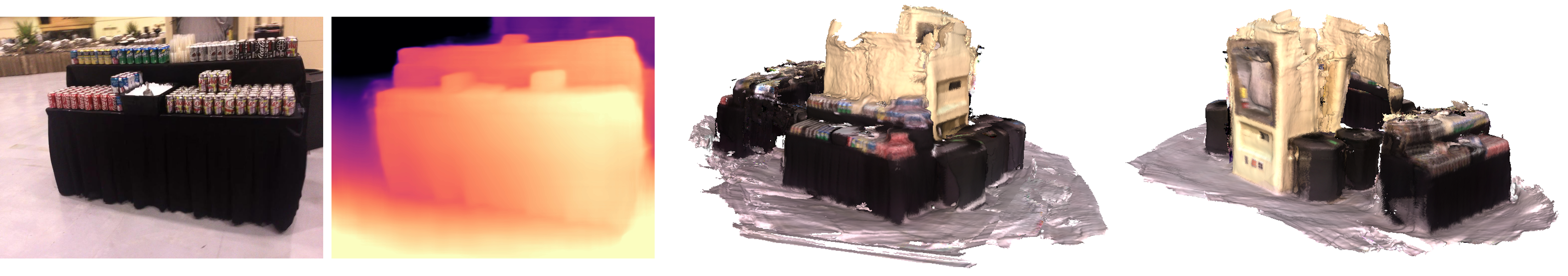}
\end{center}
\vspace{-0.5cm}
\caption{Global reconstruction on ScanNet (scene0228\_00) using the refined depth maps by GeoRefine.}
\label{fig:scannet_fusion}
\end{figure*}

\subsection{ScanNet}

ScanNet~\cite{dai2017scannet} is an indoor RGB-D dataset consisting of more than 1500 scans. This dataset was captured by a handheld device, so motion blur exists in most of the sequences, posing challenges both for monocular SLAM and depth refinement. Moreover, camera translations in this dataset are small as most of the sequences are from small rooms (\eg, bathrooms and bedrooms). To test our GeoRefine, we sample three sequences that have relatively larger camera translations and run our system using NYUv2-pretrained DPT~\cite{ranftl2021vision} as the base model. The results are summarized in Tab.~\ref{tab:scannet_eval_sup}. The pretrained DPT model performs well on ScanNet, reaching {\it Abs Rel} of 6.3\% to 8.0\%, probably due to dataset similarity between ScanNet and NYUv2. Our GeoRefine continues to improve the depth results in most of the metrics. In particular, on scene0228\_00, our system reduces {\it Abs Rel} from 8.0\% to 5.0\% and increases $\delta_1$ from 93.1\% to 97.9\%. Qualitative results can be found in Fig.~\ref{fig:scannet_fusion} and the attached video.


\begin{table}[!t]
    \centering
    \caption{Quantitative depth evaluation on ScanNet.}
    \label{tab:scannet_eval_sup}
    \resizebox{1.0\textwidth}{!}{
    \begin{tabular}{|l|c|c|c|c|c|c||c|c|c|c|c|c|}
    \hline
    \multirow{2}{*}{Method} & 
    \multicolumn{6}{c||}{Monocular} &  \multicolumn{6}{c|}{pRGBD} \\
    \cline{2-13}
    ~ & MAE $\downarrow$ & AbsRel $\downarrow$ & RMSE $\downarrow$ & $\delta_1$ $\uparrow$ & $\delta_2$ $\uparrow$ & $\delta_3$ $\uparrow$ & MAE $\downarrow$ & AbsRel $\downarrow$ & RMSE $\downarrow$  & $\delta_1$ $\uparrow$ & $\delta_2$ $\uparrow$ & $\delta_3$ $\uparrow$ \\
    \hline
    \multicolumn{13}{c}{scene0084\_00} \\
    \hline
    DPT~\cite{ranftl2021vision} &  0.118  &   0.072     &   0.164  &  0.959  &   0.994  &   {\bf 0.999} & 0.118  &   0.072     &   0.164  &  0.959  &   {\bf 0.994}  &   0.999   \\
    Ours-DPT &   {\bf 0.099}  &   {\bf 0.062}  &   {\bf 0.137}  &  {\bf 0.967}  &   0.993  &  {\bf 1.000}   &{\bf 0.089}  &   {\bf 0.052}  &   {\bf 0.145}  &  {\bf 0.983}  &  {\bf 0.995}  &  0.997 \\
    \hline
    \multicolumn{13}{c}{scene0228\_00} \\
    \hline 
    DPT~\cite{ranftl2021vision} &   0.205  &   0.080   &  0.380   &  0.931  &   0.986  &   0.998  & 0.205  &   0.080   &  0.380   &  0.931  &   0.986  &   0.998  \\
    Ours-DPT &   {\bf 0.132}  &  {\bf 0.050}   &   {\bf 0.272}  &  {\bf 0.979}  &   {\bf 0.996}  &  {\bf 0.999}  & {\bf 0.141} & {\bf 0.051} & {\bf 0.361} & {\bf 0.980} & {\bf 0.996} & {\bf 0.998} \\
    \hline
    \multicolumn{13}{c}{scene0451\_05} \\
    \hline 
    DPT~\cite{ranftl2021vision} &   0.184  &   0.080   &  0.252   &  0.947  &   {\bf 0.997}  &   {\bf 1.000}  & 0.184  &   0.080   &  0.252   &  0.947  &   {\bf 0.997}  &   {\bf 1.000}  \\
    Ours-DPT &   {\bf 0.164}  &  {\bf 0.065}   &   {\bf 0.248}  &  {\bf 0.961}  &   0.995  &  0.999  & {\bf 0.153} & {\bf 0.061} & {\bf 0.237} & {\bf 0.967} & 0.996 & 0.999 \\
    \hline
    \end{tabular}
    }
    \vspace{-0.3cm}
\end{table}

\subsection{KITTI}
\begin{table*}[!t]
       	\centering
       	\small
       	\caption{Depth evaluation results on the KITTI Eigen split test set. M: self-supervised monocular supervision; S: self-supervised stereo supervision; D: depth supervision; Align: scale alignment; Y: Yes; N: No. `-' means the result is not available from the paper. Best numbers in each block is marked in bold.}
\label{tab:kitti_tab}
\vspace{0.2cm}
\resizebox{1.0\textwidth}{!}{
\begin{tabular}{|l|l|c|c|c|c|c|c|c|c|c|}
\hline
    & \multirow{2}{*}{Method}           &   \multirow{2}{*}{Train}  &   \multirow{2}{*}{Align}      & \multicolumn{4}{c|}{Error Metric}    & \multicolumn{3}{c|}{Accuracy Metric}   \\
\cline{5-11}
 &                                &   &    & {Abs Rel} & {Sq Rel} & {RMSE}  & {RMSE log} & $\delta_1$ & $\delta_2$ & $\delta_3$ \\
 \hline
 
 \parbox[t]{3mm}{\multirow{6}{*}{\rotatebox[origin=c]{90}{Supervised }}}&Eigen~\cite{eigen2014depth}               & D&  N   & 0.203            & 1.548         & 6.307            & 0.282             & 0.702                             & 0.890                                                   & 0.890   \\
&Liu~\cite{liu2015learning}               & D&  N   & 0.201            & 1.584         & 6.471          & 0.273             & 0.680                             & 0.898                                                     & 0.967   \\
&Kuznietsov~\cite{kuznietsov2017semi}               & DS  & N  & 0.113            & 0.741         & 4.621          & 0.189             & 0.862                             & 0.960                                                     & 0.986   \\
&SVSM FT~\cite{luo2018every}               & DS  & N  & {0.094}            & {0.626}        & 4.252          & 0.177             & 0.891                             &0.965                                                     & 0.984   \\
&Guo~\cite{guo2018learning}               & DS  & N  & 0.096           & 0.641         & { 4.095}          & {0.168}             & {0.892}                            &{0.967}                                                   & {0.986}   \\
&DORN~\cite{fu2018deep}               & D  & N  & {\bf 0.072}            & {\bf 0.307}         & {\bf 2.727}          & {\bf 0.120}             &  {\bf 0.932}                             &{\bf 0.984}                                                    & {\bf 0.994}   \\
\hline
 
                                \hline
                                 \parbox[t]{4mm}{\multirow{23}{*}{\rotatebox[origin=c]{90}{ Self-Supervised}}}&
Yang~\cite{yang2017unsupervised}                            & M&  Y   & 0.182            & 1.481           & 6.501          & 0.267             & 0.725                              & 0.906                                                     & 0.963                                                     \\
&Mahjourian~\cite{mahjourian2018unsupervised}                      & M&   Y  & 0.163            & 1.240           & 6.220          & 0.250             & 0.762                              & 0.916                                                     & 0.968                                                     \\
&Klodt~\cite{klodt2018supervising}               & M&  Y   & 0.166            & 1.490         & 5.998          & -             & 0.778                             &0.919                                                     & 0.966   \\
&DDVO~\cite{wang2018learning}                            & M&  Y   & 0.151            & 1.257           & 5.583          & 0.228             & 0.810                              & 0.936                                                     & 0.974                                                     \\
&GeoNet~\cite{yin2018geonet}                          & M&  Y   & 0.149            & 1.060           & 5.567          & 0.226             & 0.796                              & 0.935                                                     & 0.975                                                     \\
&DF-Net~\cite{zou2018dfnet}                          & M&  Y   & 0.150            & 1.124           & 5.507          & 0.223             & 0.806                              & 0.933                                                     & 0.973                                                     \\
&Ranjan~\cite{ranjan2019competitive}                          & M&  Y   & 0.148            & 1.149           & 5.464          & 0.226             & 0.815                              & 0.935                                                     & 0.973                                                     \\
&EPC++~\cite{luo2018every}                           & M&  Y   & 0.141            & 1.029           & 5.350          & 0.216             & 0.816                              & 0.941                                                     & 0.976                                                     \\
&Struct2depth(M)~\cite{casser2019depth}                & M&  Y   & 0.141            & 1.026           & 5.291          & 0.215             & 0.816                              & 0.945                                                     & 0.979                                               \\
&WBAF~~\cite{zhou2020windowed}                            & M& Y &             0.135 &    0.992 &          5.288 &        0.211 &   0.831 &  0.942 &  0.976 \\
&pRGBD-Refined~\cite{tiwari2020pseudo}          & M&  Y   & 0.113   & 0.793  & 4.655 & {\bf 0.188}    & {0.874}                        & 0.960                                            & 0.983   \\
&Luo~\cite{luo2020consistent} & M& Y & 0.130 & 2.086 &  4.876 & 0.205 & 0.878 & 0.946 & 0.970 \\
&Li~\cite{li2021generalizing} & M& Y& 0.106 & {\bf 0.701} & {\bf 4.129} & 0.210 & 0.889 & {\bf 0.967} & {\bf 0.984} \\

\cline{2-11}
&Garg~\cite{garg2016unsupervised}                            & S&  N   & 0.152            & 1.226           & 5.849          & 0.246             & 0.784                              & 0.921                                                     & 0.967                                                             \\
&3Net (R50)~\cite{poggi2018learning}                      & S&  N   & 0.129            & 0.996           & 5.281          & 0.223             & 0.831                              & 0.939                                                     & 0.974                                                     \\
&Monodepth2-S~\cite{godard2019digging}                & S&  N   & {0.109}            & 0.873           & {4.960}         & 0.209             & {0.864}                              & {0.948}                                                    & 0.975    \\
&SuperDepth ~\cite{pillai2019superdepth}  & S&  N   & 0.112            & 0.875           & {4.958}          & {0.207}            & {0.852}                           & {0.947}                                                     & {0.977}                                                                                        \\
&monoResMatch~~\cite{tosi2019learning}            & S & N &             0.111 &   0.867  &         4.714  &       0.199 &   0.864 &  {0.954} &  {0.979}\\
&DepthHints~~\cite{watson2019self}                          & S&   N    &       {0.106} &   {0.780}   &        {4.695}   &      {0.193}  &  {0.875} &  {\bf 0.958} &  {\bf 0.980}\\
&DVSO~\cite{yang2018deep}               & S&  N   & {\bf 0.097}            & {\bf 0.734}         & {\bf 4.442}         & {\bf 0.187}             & {\bf 0.888}                             &{\bf 0.958}                                                     & {\bf 0.980}    \\
\cline{2-11}
&UnDeepVO~~\cite{li2018undeepvo}               & MS &  N  & 0.183            & 1.730         & 6.570          & 0.268             & -                             &-                                                     & -   \\
&EPC++~~\cite{luo2018every} & MS &  N  & {0.128}            & {0.935}         & {5.011}          & {0.209}           & {0.831}                             &{0.945}                                                    &  {0.979}   \\
&Monodepth2~\cite{godard2019digging}                & MS  & N  &  {0.106}            & {0.818}           & {4.750}          & {0.196}             & {0.874}                              & {0.957}                                                     & {0.979}  \\

&Ours-MD2-Mono & (S)M & Y & \textbf{0.096} & 0.766 & 4.436 & {\bf 0.177} & \textbf{0.902} & {\bf 0.963} &
    {\bf 0.982} \\
\hline
\end{tabular}
}
\end{table*}

We show the depth results on KITTI in Tab.~\ref{tab:kitti_tab}. The motion threshold for keyframes (or per-frame) is set to 0.25 m (or 0.05 m), $\lambda_m$ to 0.01, and three frames (\ie, 0, -1, 1) are used to build the loss; other parameters remain the same as in the main paper. 
Compared to the base model Monodepth2, our GeoRefine reduces {\it Abs Rel} by 1\% and improves $\delta_1$ by 2.8\%. However, due to moving objects in KITTI, the improvement by our system is not as significant as in non-dynamic indoor environments.

\subsection{Runtime}
Our RAFT-SLAM and online dense mapping modules run in parallel with a rough $1$ fps runtime in total. On the RAFT-SLAM side, since we only publish one pair image each time to the RAFT network end in a down-scaled resolution, the per-frame tracking can be executed at $5$ fps. For dense mapping, the per-frame refinement step runs efficiently with around $10$ fps when using the pretrained Monodepth2 model in a lower resolution and or using the pretrained DPT model. Keyframe refinement is the most time-consuming step in our system, costing around 300 ms each time. The rest of runtime is consumed by data loading, pre-processing, and cross-module communication, which can be further optimized in a future version.

%
%
\bibliographystyle{splncs04}
\bibliography{georefine}
\end{document}